%% file: main.tex
\documentclass[10pt,journal,compsoc]{IEEEtran}

\usepackage[utf8]{inputenc} 
\usepackage[T1]{fontenc}    
\usepackage{url}            
\usepackage{booktabs}       
\usepackage{amsfonts}       
\usepackage{amsmath}
\usepackage{nicefrac}       
\usepackage{microtype}      
\usepackage{amsthm}
\usepackage{amssymb}
\usepackage{rotating}
\usepackage{booktabs}
\usepackage{graphicx}
\usepackage{subfigure}
\usepackage{caption}
\usepackage{wrapfig}
\usepackage{bm}
\usepackage{multirow}
\usepackage{mathtools}
\usepackage{lscape}
\usepackage{hyperref}
\usepackage{url}
\usepackage{bbm}
\newcommand{\eat}[1]{}

\newtheorem{definition}{Definition}

\newtheorem{remark}{Remark}
\input{math_commands.tex}

\usepackage{color,colortbl}
\definecolor{ballblue}{rgb}{0.13, 0.67, 0.80}
\definecolor{amaranth}{rgb}{0.90, 0.17, 0.31}
\definecolor{olive}{rgb}{0.5, 0.5, 0.0}
\definecolor{Gray}{gray}{0.85}
\newcolumntype{a}{>{\columncolor{Gray}}c}
\definecolor{blush}{rgb}{0.87, 0.36, 0.51}

\ifCLASSOPTIONcompsoc
  \usepackage[nocompress]{cite}
\else
  \usepackage{cite}
\fi

\ifCLASSINFOpdf
\else
\fi

\begin{document}

\title{Relation Matters: Foreground-aware Graph-based Relational Reasoning for Domain Adaptive Object Detection}

\author{Chaoqi~Chen,
	Jiongcheng~Li,
	Hong-Yu Zhou,
	Xiaoguang Han,
	Yue~Huang,
	Xinghao~Ding,
	and~Yizhou~Yu,~\IEEEmembership{Fellow,~IEEE}
	\IEEEcompsocitemizethanks{
		\IEEEcompsocthanksitem This work was partially funded by National Natural Science Foundation of China under Grants U19B2031 and 82172033. (Corresponding authors: Yizhou Yu and Yue Huang.)
		\IEEEcompsocthanksitem Chaoqi Chen, Hong-Yu Zhou, and Yizhou Yu are with the Department of Computer Science, The University of Hong Kong, Hong Kong. Email: \{cqchen1994, whuzhouhongyu\}@gmail.com, yizhouy@acm.org. 
		\IEEEcompsocthanksitem Jiongcheng Li, Yue Huang, and Xinghao Ding are with the Department of Information and Communication Engineering, Xiamen University, Xiamen, China. Email: jiongchengli@stu.xmu.edu.cn,  \{yhuang2010, dxh\}@xmu.edu.cn.
		\IEEEcompsocthanksitem Xiaoguang Han is with the Shenzhen Research Institute of Big Data, The Chinese University of Hong Kong (Shenzhen), Shenzhen, China. Email: hanxiaoguang@cuhk.edu.cn.
		}}

\markboth{IEEE Transactions on Pattern Analysis and Machine Intelligence}%
{Shell \MakeLowercase{\textit{et al.}}: Bare Demo of IEEEtran.cls for Computer Society Journals}

\IEEEtitleabstractindextext{%
\begin{abstract}
Domain Adaptive Object Detection (DAOD) focuses on improving the generalization ability of object detectors via knowledge transfer.
Recent advances in DAOD strive to change the emphasis of the adaptation process from global to local in virtue of fine-grained feature alignment methods. 
However, both the global and local alignment approaches fail to capture the topological relations among different foreground objects as the explicit dependencies and interactions between and within domains are neglected. 
In this case, only seeking one-vs-one alignment does not necessarily ensure the precise knowledge transfer.
Moreover, conventional alignment-based approaches may be vulnerable to catastrophic overfitting regarding those less transferable regions (e.g. backgrounds) due to the accumulation of inaccurate localization results in the target domain.
To remedy these issues, we first formulate DAOD as an open-set domain adaptation problem, in which the foregrounds and backgrounds are seen as the ``known classes'' and ``unknown class'' respectively. 
Accordingly, we propose a new and general framework for DAOD, named Foreground-aware Graph-based Relational Reasoning (FGRR), which incorporates graph structures into the detection pipeline to explicitly model the intra- and inter-domain foreground object relations on both pixel and semantic spaces, thereby endowing the DAOD model with the capability of relational reasoning beyond the popular alignment-based paradigm. 
FGRR first identifies the foreground pixels and regions by searching reliable correspondence and cross-domain similarity regularization respectively.
The inter-domain visual and semantic correlations are hierarchically modeled via bipartite graph structures, and the intra-domain relations are encoded via graph attention mechanisms. 
Through message-passing, each node aggregates semantic and contextual information from the same and opposite domain to substantially enhance its expressive power.
Empirical results demonstrate that the proposed FGRR exceeds the state-of-the-art performance on four DAOD benchmarks.  
\end{abstract}

\begin{IEEEkeywords}
Domain adaptive object detection, foreground-aware, relational reasoning, graph structure, intra- and inter-domain.
\end{IEEEkeywords}}

\maketitle

\IEEEdisplaynontitleabstractindextext

\IEEEpeerreviewmaketitle
\IEEEraisesectionheading{\section{Introduction}\label{sec:introduction}}
\IEEEPARstart{A}{s} one of the most fundamental problems in computer vision, object detection has gained a great surge of development in the past decade, 
owing to the renaissance in deep learning and the explosive increase of labeled training data. 
Nevertheless, the impressive performance gains rely on a strong pre-assumption that the training and test data are drawn from identical distribution, which is challenged to be satisfied in many real-world applications, such as autonomous driving and medical image analysis. 
Moreover, manually collecting large-scale instance-level annotated data in various domains is labor-intensive and impractical. 
An intuitive solution is to directly apply the off-the-shelf object detection models trained on the source domain to a new target domain.     
However, domain shift~\cite{quionero2009dataset,torralba2011unbiased} hinders the deployment of models and emerges as an inevitable challenge.
Such a dilemma has inspired the research on Unsupervised Domain Adaptation~(UDA)~\cite{pan2010survey}, which aims at mitigating the domain disparity via transferring knowledge from a label-rich domain to a new fully unlabeled domain.

The mainstream paradigm for UDA is to learn domain-invariant features by minimizing the discrepancy of source and target feature distributions. 
Current UDA approaches can be roughly divided into two categories: 1) statistics matching, which targets on aligning feature representations across domains in virtue of statistical distribution divergence~\cite{gong2012geodesic,fernando2013unsupervised,long2015learning,long2017deep,zellinger2017central,peng2019moment}; 2) adversarial learning, which adversarially learns domain-invariant representations based on the two-player game between domain discriminator and feature
extractor~\cite{ganin2015unsupervised,tzeng2017adversarial,long2018conditional,xie2018learning,Chen_2019_CVPR,jiang2020implicit}.
By doing so, UDA has shown remarkable progress in image classification and semantic segmentation problems. 

Compared to the conventional problems of UDA, object detection is a more challenging problem as it requires one to simultaneously achieve adaptation under the classification and regression settings. 
In this paper, our objective is to investigate the UDA techniques for object detection, namely, Domain Adaptive Object Detection~(DAOD).
In light of the local nature of detection tasks, state-of-the-art DAOD approaches~\cite{chen2018domain,zhu2019adapting,saito2019strong,cai2019exploring,He_2019_ICCV,chen2020harmonizing,xu2020cross,zheng2020cross,xu2020exploring,hsu2020every,wu2021instance} are dedicated to change the focus of adaptation process from global to local by means of fine-grained feature alignment methods regarding the foreground objects. Typically, they incorporate the adversarial feature training within de facto detection frameworks~\cite{ren2015faster,liu2016ssd} 
to form the baseline model. 
Then, elaborate local alignment modules, such as mining foreground regions via attention-like mechanism~\cite{chen2020harmonizing,zheng2020cross,hsu2020every}, are imposed on the baseline model to extract and align the discriminative features at different feature levels, \emph{i.e.,} local-level, global-level, and instance-level.

\begin{figure}[!t]
	\centering
	\subfigure[]{\label{fig1:a}\includegraphics[width=0.45\textwidth]{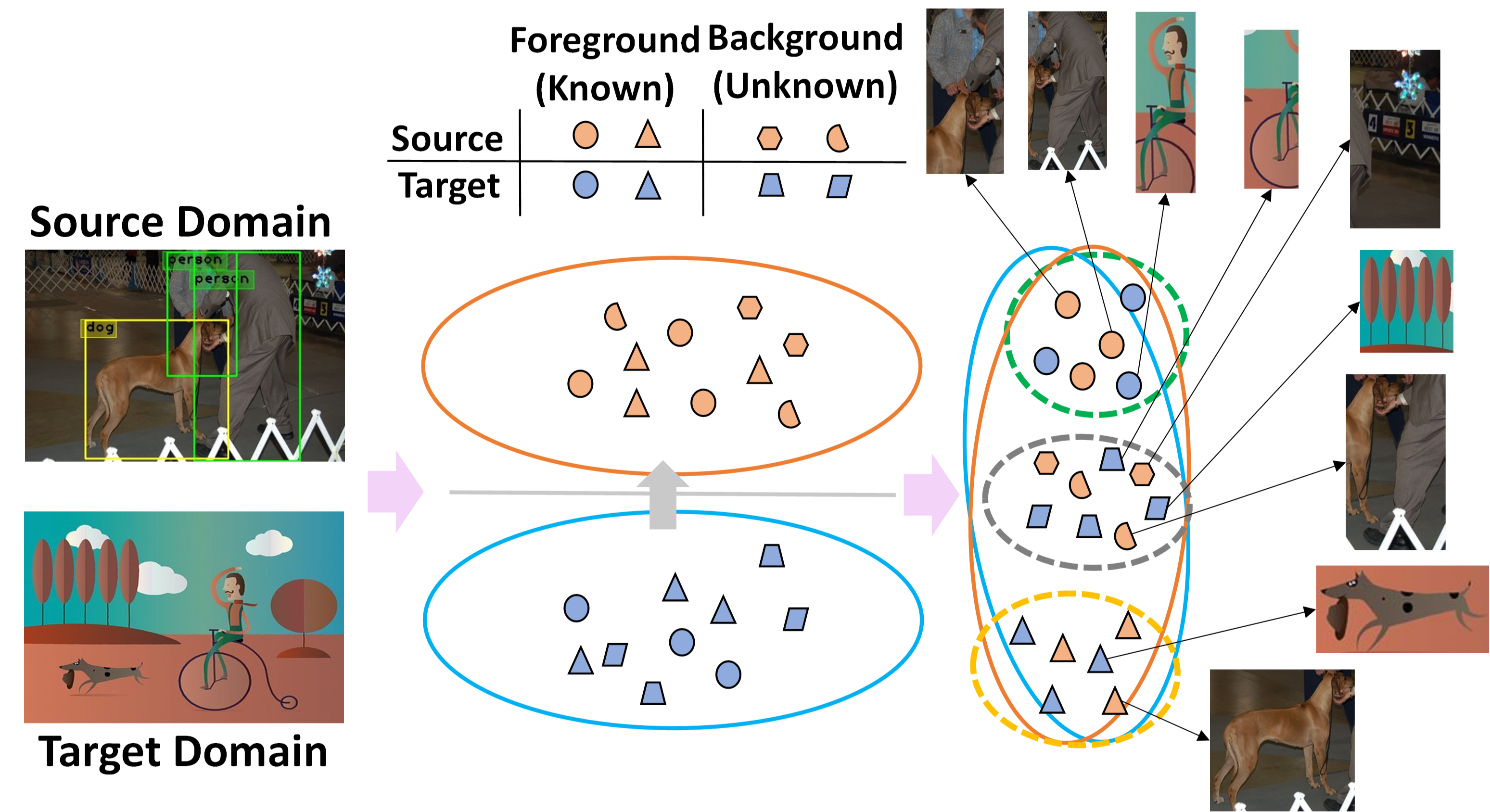}}
	\subfigure[]{\label{fig1:b}\includegraphics[width=0.45\textwidth]{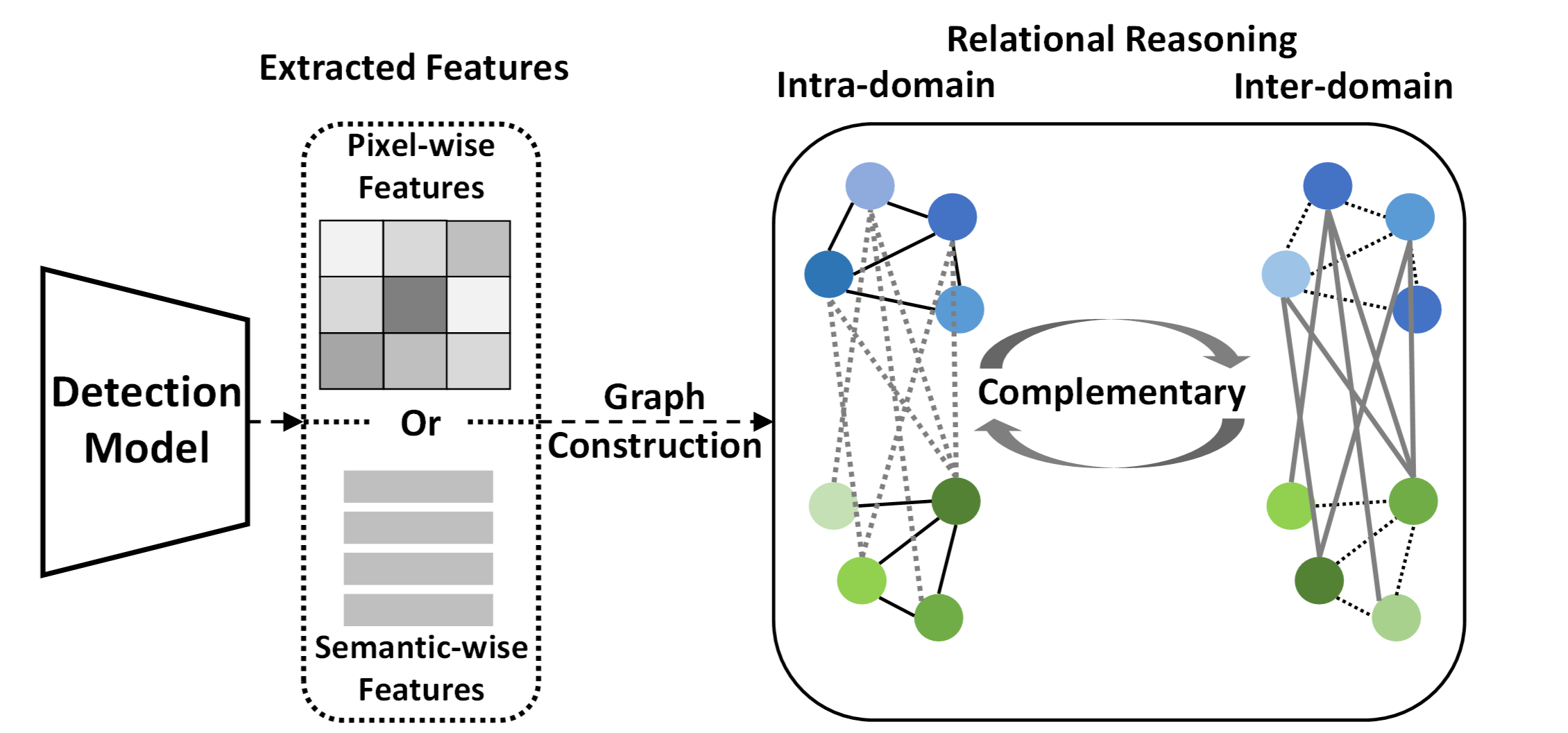}}
	\caption{(a) We formulate DAOD as an open-set domain adaptation problem, where foreground is the known classes and background is the unknown class. (b) Our method reasons the foreground object relationships in both intra- and inter-domain. The graph construction process mines the foreground pixels or regions based on the extracted features.}
	\vspace{-0.3cm}
\end{figure}

However, both the domain and region alignment approaches in the literature, such as weak global alignment~\cite{saito2019strong}, multi-adversarial alignment~\cite{He_2019_ICCV}, feature calibration~\cite{chen2020harmonizing}, and coarse-to-fine alignment~\cite{zheng2020cross}, fail to capture the topological relations among foreground objects during the adaptation process.
In this regard, simply seeking explicit one-vs-one matching no matter in image-level or instance-level cannot ensure the precise knowledge transfer since they do not model the explicit dependencies and interactions between and within domains. 
Moreover, such strict alignment may be prone to result in catastrophic overfitting regarding those less transferable regions (e.g. backgrounds) due to the accumulation of inaccurate localization results in the target domain. Consequently, existing DAOD methods may be sensitive to the size of domain discrepancy, \emph{i.e.,} when the domain discrepancy is large, the unstable alignment process will significantly affect the adaptation performance. 
How to explore the relationships among different foreground objects is vital for a holistic understanding of the scene in each domain and the precise knowledge transfer between domains, but remains out of reach for current approaches. 


Remedying these issues, we first propose to formulate DAOD as an Open Set Domain Adaptation (OSDA) problem~\cite{panareda2017open}. 
Unlike the closed set domain adaptation, wherein the source and target domains contain the same classes, 
OSDA should explicitly identify and isolate the unknown class before reducing the distributional disparity of known classes across domains.  
In the task of DAOD, we argue that both the so-called known and unknown classes refer to the category of a pixel or region rather than the whole image.
As shown in Fig.~\ref{fig1:a}, we observe that the backgrounds are distinct (non-transferable) across domains (\emph{e.g.,} ``billboard'' in the source domain and ``tree'' in the target domain) and can be regarded as unknown class, while the foregrounds possess more common features across domains (\emph{e.g.,} ``dog'' in both domains has four legs and one head) and can be regarded as known class. 
Compared to classification based OSDA, the decision boundaries among features of target known and unknown classes in the detection pipeline are ambiguous and these features may be highly entangled, impeding the DAOD model to learn object-wise invariant representations across domains.
This motivates us to design DAOD algorithms in the following two steps: 
(1) Make a distinction between foreground and background feature representations in an unsupervised manner. 
(2) Perform relational reasoning to model the foreground object interactions and align the corresponding categories in both domains. 

Grounded on these findings, we propose a Foreground-aware Graph-based Relational Reasoning (FGRR) framework for DAOD to explicitly model the intra- and inter-domain foreground object interactions on both visual (pixel-level) and semantic spaces, thereby endowing the detection model with the capability of relational reasoning beyond the mainstream one-vs-one feature alignment paradigm. 
To be specific, FGRR consists of two key components, pixel-level and semantic-level relational reasoning (cf. Figure~\ref{fig1:b}).
For pixel-level relational reasoning, we search pixel-wise correspondence to find out foreground pixels via mutual nearest neighbor constraint.
The searched pixels is the graph nodes, and then we model the inter-domain relations via bipartite graph learning and the intra-domain relations via graph attention mechanism. Through message-passing and feature aggregation, each foreground node can be aware of the contextual information within domain and the topological interactions across domains.  
For semantic-level relational reasoning, we discriminate the target instance-level features to define graph nodes by choosing proposals with high-confidence prediction. When reasoning in the high-level semantic space, we devise a cross-domain similarity regularization strategy, which penalizes the ones of nodes that are more likely to be backgrounds and enhance the connections between foreground nodes, to further identify and isolate the backgrounds so as to construct semantic-level bipartite graph. In addition, the intra-domain graph edges are characterized from two aspects: geometric constraints and semantic similarities. By doing so, the semantic correlations are propagated through nodes methodically, helping the DAOD model understand the high-level semantic concepts and reason their relationships. 






Noting that the proposed FGRR embraces the complementary strengths of the inter- and intra-domain reasoning modules. On the one hand, we hierarchically learn the relational structures between domains via the proposed inter-domain graph reasoning modules, which helps the detection model capture more informative commonalities for constructing intra-domain relation graphs. On the other hand, we use intra-domain relation graphs to abstract away from dense feature maps and model the intra-domain foreground pixel/object interactions, which provide high-quality relational structures for inter-domain bipartite graph learning. In addition, upon observing that the cross-domain object-level variations based on the image-level features are prone to result in negative transfer but difficult to capture, we develop an object-aware reweighting mechanism to regularize image-level adaptation by explicitly accounting for the object-level variations from two complementary aspects. 

The contributions of our work are summarized as follows: 
\begin{itemize}
	\item We formulate DAOD as an OSDA problem, which has not been explored by the literature and provides a new and general perspective to mitigate the domain shift for adapting object detectors. 
	\item We make the first attempt to reason the foreground object relationships and interactions via graph-based structures rather than following the mainstream one-vs-one feature alignment paradigm.  
	\item We propose a novel DAOD framework, named FGRR, which seamlessly incorporates the inter- and intra-domain relational reasoning modules into the detection pipeline. 
	These modules are imposed to the pixel-wise and instance-wise features to jointly learn visual and semantic relationships. 
	\item We conduct extensive experiments based on modern object detectors. 
	Experimental results reveal that our FGRR models have achieved state-of-the-art results on four DAOD benchmarks: real-to-artistic, normal-to-foggy, and synthetic-to-real datasets, as well as cross-site mammogram mass detection (from public to in-house datasets).
\end{itemize}

A preliminary version of this paper was presented in~\cite{chen2021dual}. We have made substantial changes to that conference paper. 
\textbf{(1)} Besides modeling the inter-domain relations, our FGRR further explores the complementary effect of intra-domain relations to help the source and target graph nodes attend and reason over their intra-domain neighborhoods' features instead of treating each pixel or region separately in each domain, thereby capturing the internal object structures and contextual information.
\textbf{(2)} An image-level object-aware reweighting mechanism is introduced to deal with the object-level variations based on image-level features, which are difficult to be estimated since the object information here is implicit and ambiguous. 
This mechanism is simple yet effective by only utilizing the searched foreground pixels to compute the weight of input image pairs during image-level adaptation. 
\textbf{(3)} We improve the mechanisms of defining graph nodes to find out more reliable foreground pixels and regions.
\textbf{(4)} We additionally conduct experiments on two widely-used DAOD benchmarks (normal-to-foggy and synthetic-to-real) to comprehensively evaluate the effectiveness of the proposed FGRR. In addition, we also evaluate our FGRR on a practical application, \emph{i.e.,} cross-site mammogram mass detection. More ablation studies and analyses are provided to demonstrate the contributions of newly proposed components.  
\textbf{(5)} We provide additional insights on relational reasoning for DAOD, more complete introduction and analysis, as well as more elaborated literature review and technical details regarding the proposed method.  

\section{Related Works}
\label{sec:related_work}
\subsection{Unsupervised Domain Adaptation~(UDA)}
In the literature, UDA has achieved remarkable success for bridging the domain disparity between two different distributions in image classification, semantic segmentation, and object detection problems.
Specifically, a typical solution for UDA is to align the source and target feature representations in the shared latent space by incorporating well-defined divergence measures into deep architectures.
For example, DDC~\cite{tzeng2014deep} and DAN~\cite{long2015learning} resort to Maximum Mean Discrepancy (MMD)~\cite{gretton2012kernel} for measuring the domain discrepancy in the task-specific layers of deep networks. 
Correlation Alignment (CORAL)~\cite{sun2016deep} aligns the second-order statistics of source and target distributions. Central Moment Discrepancy (CMD)~\cite{zellinger2017central} matches higher order central moments of probability distributions via the means of order-wise moment differences. Margin Disparity Discrepancy (MDD)~\cite{zhang2019bridging}, which characterizes the disagreements of multi-class scoring hypotheses, extends previous UDA theories to the case of multiclass classification. 

Another line of research was inspired by the essence of Generative Adversarial Nets (GAN)~\cite{goodfellow2014generative}. DANN~\cite{ganin2016domain} presents a domain-adversarial training strategy to learn domain-invariant representations by adversarially confusing a domain discriminator with the help of a Gradient Reversal Layer (GRL), and this work motivates a number of follow-up methods due to its strong efficacy and scalability~\cite{xie2018learning,zhang2019domain,Chen_2019_CVPR,chen2019transferability,wang2020self}. 
For example, ADDA~\cite{tzeng2017adversarial} improves DANN by separating the source and target feature extractors to learn target representations in an asymmetric way. Saito \textit{et al.}~\cite{saito2018maximum} utilizes two task-specific classifiers as discriminators, and a feature extractor learns to fool the classifiers.
CDAN~\cite{long2018conditional} conditions the adversarial training on categorical information conveyed in the label classifier. 
In addition, the works of~\cite{bousmalis2017unsupervised,liu2017unsupervised,russo2018source,hu2018duplex,hoffman2018cycada,Tran_2019_CVPR} directly utilize the GAN-based image-to-image translation techniques, e.g., CycleGAN~\cite{zhu2017unpaired}, to generate source-like target images or/and target-like source images to achieve pixel-level adaptation. Other representative approaches to UDA include self-ensembling~\cite{french2018self}, similarity learning~\cite{pinheiro2018unsupervised}, pseudo-labeling~\cite{saito2017asymmetric,zou2018unsupervised,zou2019confidence}, structurally regularized deep clustering~\cite{tang2020unsupervised}, to name a few.

Particularly, several UDA works attempt to explore intrinsic characteristics of source and target data in addition to domain and category label information.
To be specific, Ma~\emph{et al.}~\cite{ma2019gcan} incorporate GCN into the adaptation pipeline to exploit the data structure for bridging source and target domains.
Luo~\emph{et al.}~\cite{luo2020adversarial} introduce an adversarial bipartite graph learning algorithm to model the source-target interactions for video-based UDA. Kang~\emph{et al.}~\cite{kang2020pixel} delve into the pixel-wise one-to-one association problem for domain adaptive semantic segmentation. 
These methods, however, fail to explore the topological correspondence across domains and thus cannot endow the UDA model with the capability of cross-domain relation reasoning.   
More importantly, existing UDA methods focus on the closed set adaptation and cannot generalize to OSDA~\cite{panareda2017open} scenarios. 
Moreover, state-of-the-art OSDA methods~\cite{liu2019separate,baktashmotlagh2019learning,pan2020exploring,luo2020progressive} are tailored for classification problems and cannot be applied to solve the detection tasks, 
where the foreground objects and backgrounds may come from the same image and can be naturally regarded as the known and unknown classes.  

\subsection{Domain Adaptive Object Detection~(DAOD)}
\subsubsection{Object Detection}
Object detection, one of the most longstanding and challenging problems in computer vision, aims at localizing and classifying all object instances of given categories in natural images~\cite{liu2020deep}.
The current state-of-the-art is shaped by deep learning based approaches. 
This paper focuses on exploring the adaptability of modern object detectors, hence we briefly review a few representative object detectors. 
The series of region-based convolutional networks, such as R-CNN~\cite{girshick2014rich}, Fast R-CNN~\cite{girshick2015fast}, and Faster R-CNN~\cite{ren2015faster}, have demonstrated strong capability in achieving higher detection precision. 
These methods first get a set of object proposal candidates generated by the region proposal mechanisms, and then the proposals are further classified and their locations are refined by regression in the second stage. 
The other mainstream pipeline is one-stage detector, such as SSD~\cite{liu2016ssd}, YOLO~\cite{redmon2016you,redmon2017yolo9000}, and RetinaNet~\cite{lin2017focal}, which directly conducts the category confidence prediction and the bounding box regression in a single-shot manner. 
By doing so, these detectors have shown a clear superiority in terms of inference speed.
In addition, anchor-free one-stage object detectors, such as CornerNet~\cite{law2018cornernet} and FCOS~\cite{tian2019fcos}, eliminate the need for anchor boxes and thus result in a much simpler training process. 

\subsubsection{Domain Adaptive Object Detection (DAOD)}
DAOD~\cite{li2020deep,oza2021unsupervised} aims to mitigate the distributional shift problem in object detection.
Domain Adaptive Faster R-CNN~\cite{chen2018domain} pioneers this line of research by adversarially learning domain-invariant representations on both image-level and instance-level within the Faster R-CNN framework. 
In view of the local nature of object detection task, most existing DAOD methods~\cite{zhu2019adapting,saito2019strong,cai2019exploring,He_2019_ICCV,chen2020harmonizing,xu2020cross,zheng2020cross,xu2020exploring,hsu2020every,su2020adapting,zhao2020collaborative,sindagi2020prior,vs2021mega,wu2021instance,deng2021unbiased,wu2021vector} resort to learning the local feature patterns in the target domain in virtue of elaborate fine-grained feature alignment modules.

Specifically, Zhu~\emph{et al.}~\cite{zhu2019adapting} design a region mining strategy to identify the discriminative regions, and then utilize the source region proposals to reweight the target ones to induce better local alignment; 
Xu~\emph{et al.}~\cite{xu2020cross} and Zheng~\emph{et al.}~\cite{zheng2020cross} rely on the cross-domain prototype alignment~\cite{xie2018learning,Chen_2019_CVPR} to align the foreground objects with the same class label across domains. 
He~\emph{et al.}~\cite{he2020domain} propose an asymmetric tri-way Faster-RCNN to solve the labeling unfairness between domains.
Wu~\emph{et al.}~\cite{wu2021instance} explicitly extract domain-invariant instance-level features based on a progressive disentangled mechanism. Wang~\emph{et al.}~\cite{wang2021afan} and Chen~\emph{et al.}~\cite{chen2021scale} additionally incorporate the object scale into adversarial training networks.
Hsu~\emph{et al.}~\cite{hsu2020every} and VS~\emph{et al.}~\cite{vs2021mega} introduce specific networks to generate foreground-aware attention maps, which are used to achieve category-wise adversarial feature alignment. 
Although our work have the similar motivation about creating foreground-background awareness compared to some of local alignment methods, these prior efforts still focus on pairwise alignment and usually require additional attention modules.

In addition to the series of works that adapt Faster R-CNN, 
Kim~\emph{et al.}~\cite{kim2019self} aims to adapt SSD via weak self-training and adversarial background score regularization. 
DA-DETR~\cite{zhang2021detr} develops a hybrid attention module with a domain discriminator on the top of Deformable DETR~\cite{zhu2021deformable} to highlight and align the foreground features in adversarial training.
Transformer-based methods~\cite{vaswani2017attention} globally model the long-range dependencies among all encoded features while our FGRR locally models the relations based on the constructed graphs (a sparse version of encoded features). Here, the difference between global and local is that whether the self-attention operation is applied to all encoded features. 

Despite their strong efficacy on adapting certain detectors, state-of-the-art methods cannot be readily applied to distinct detection pipelines and thus fail to establish a general DAOD approach. More importantly, 
how to effectively perform inter- and intra-domain relational reasoning to capture the topological relationships among foreground objects is vital for enhancing the discriminability of feature representations in DAOD but remains the boundary to explore.

\vspace{-5pt}
\section{Methodology}
\label{sec:method}
\begin{figure*}[!t]
	\centering
	\includegraphics[width=1\textwidth]{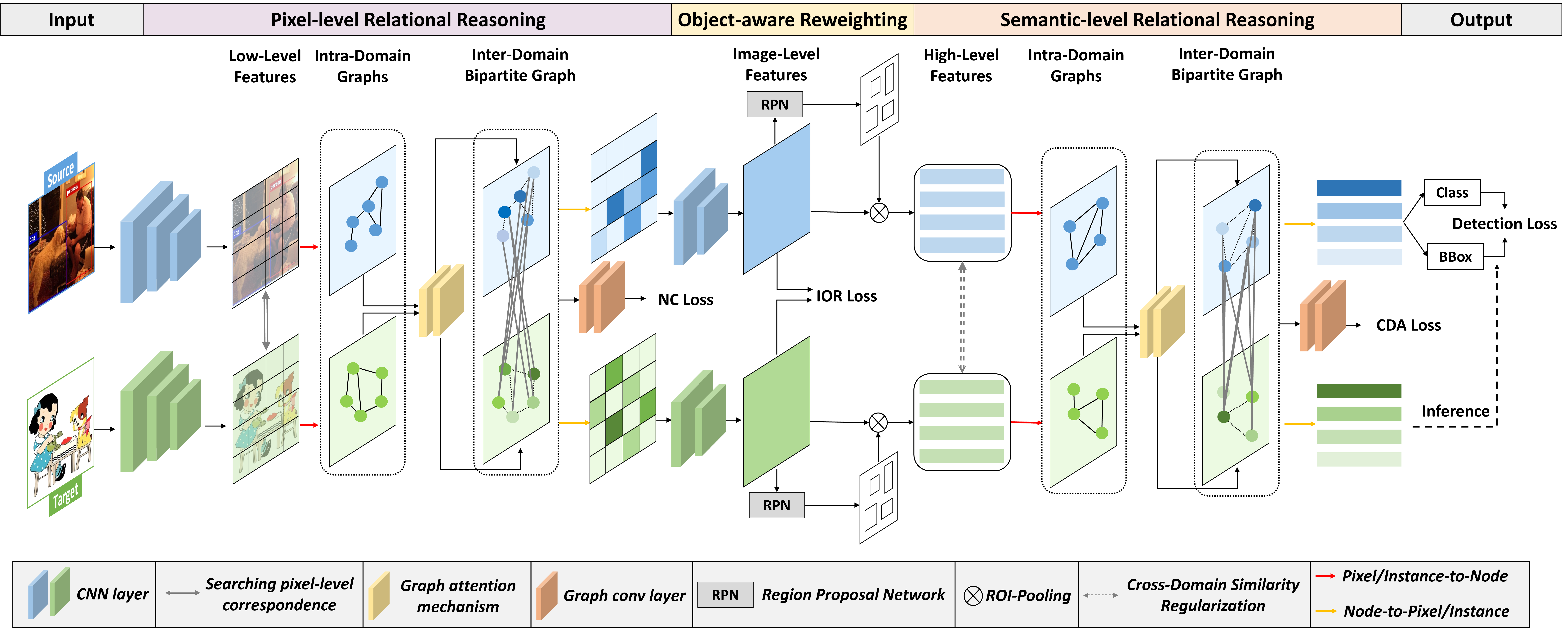}
	\caption{The overall structure of the proposed FGRR, which mainly includes the pixel-level and semantic-level relational reasoning modules as well as the image-level object-aware reweighting. NC, IOR, and CDA denote the node classification, object-aware reweighting, and category-aware domain alignment losses. 
	}\label{fig2}
\end{figure*}


In DAOD, we have access to a source domain $\mathcal{D}_s=\{({x_i^s},{y_i^s},{b_i^s})\}_{i=1}^{N_s}$ (${y_i^s}\in{\mathcal{R}^{k\times{1}}}$, $b_i^s\in\mathcal{R}^{k\times{4}}$) of $N_s$ labeled samples, and a target domain $\mathcal{D}_t \! = \! \{{x_j^t}\}_{j=1}^{N_t}$ of $N_t$ unlabeled samples. $\mathcal{D}_s$ and $\mathcal{D}_t$ are sampled from distinct data distributions, but share an identical label space ($K$ classes in all). The objective of DAOD is to transfer knowledge from $\mathcal{D}_s$ to $\mathcal{D}_t$ for improving the detection performance in $\mathcal{D}_t$. 

The overall architecture of FGRR is illustrated in Figure~\ref{fig2}, which involves three modules: pixel-level relational reasoning (low-level features), semantic-level relational reasoning (high-level features), and image-level object-aware reweighting (image-level features). 
FGRR targets on endowing the DAOD model with inter- and intra-domain relational reasoning ability.
(1) We filter out source background pixels within the bounding boxes and search target foreground pixels via mutual nearest neighbor constraint (Sec.~\ref{sec:3-1-1}).  
The searched pixels is represented as the graph nodes, and the inter- and intra-domain relations are learned via bipartite graph learning (Sec.~\ref{sec:3-1-2}) and graph attention mechanisms (Sec.~\ref{sec:3-1-3}). 
(2) When reasoning over relations between high-level visual concepts, we discriminate the target instance-level features to define graph nodes (Sec.~\ref{sec:3-2-1}), and then a cross-domain similarity regularization strategy (Sec.~\ref{sec:3-2-2}) is introduced to identify and isolate the backgrounds for constructing the semantic-level bipartite graph. 
In addition, the intra-domain graph edges are characterized from two aspects: geometric constraints and semantic similarities (Sec.~\ref{sec:3-2-3}).
(3) The object-aware reweighting (Sec.~\ref{sec:3-3}) is introduced to capture the object-level variations by learning appropriate importance weights for different input images.
The \emph{motivation} and \emph{technical details} of each proposed module are successively demonstrated in this section.

\subsection{Pixel-Level Relational Reasoning} 
Deep features in the standard CNNs must eventually transition from general to specific 
along the network~\cite{yosinski2014transferable}. 
Low-level features, obtained from the shallow layers of CNNs, have less semantics but with rich local details, \emph{e.g.,} have more clear edges and corners.
For low-level features, state-of-the-art DAOD methods to date mostly resort to global domain alignment via adversarial feature adaptation~\cite{chen2018domain,saito2019strong,He_2019_ICCV} or trying to coarsely derive the foreground regions via 
attention-related strategies~\cite{chen2020harmonizing,zheng2020cross,hsu2020every}, such as class activation maps (CAMs)~\cite{zhou2016learning} and the entropy of domain discriminator outputs~\cite{chen2020harmonizing}. 
Despite their efficacy for significantly reducing the domain disparity, current DAOD approaches for adapting low-level features still suffer from three key challenges. 
First, they cannot capture the cross-domain foreground object relations (such as object co-occurrence), and thus fail to enable fine-grained and more precise knowledge transfer. 
Second, globally matching features between domains is error-prone since the extracted features from shallow layers may be dominated by background noise,
\emph{i.e.,} the foreground and background features are highly entangled. 
Third, relations between distant foreground regions on the feature map, which are crucial for depicting intra-domain topological structures, cannot be captured on the shallow layers with small receptive field in conventional DAOD pipelines.  

To tackle the aforementioned challenges, we propose to endow the shallow layers of a DAOD model with relational reasoning ability to achieve a holistic understanding of foreground objects with the following two steps. First, we need to find out the discriminative foreground pixels by searching reliable pixel-level correspondence between domains. Second, based on the selected foreground pixels, we propose to model and reason over these pixels in both inter-domain and intra-domain by harnessing two complementary graph-based relational reasoning modules.

\subsubsection{Searching Reliable Pixel-Level Correspondence}
\label{sec:3-1-1}
Assume that we are given the source and target 3D feature maps $F_s, F_t \in \mathbb{R}^{C\times{H}\times{W}}$ extracted from shallow layer of the backbone network, \emph{i.e.,} the extracted features are low-level features. 
First of all, we aim to find source foreground pixels by using source annotations, including category labels and bounding boxes.
In object detection, the semantic label of a certain source pixel is determined by the scope of its bounding box, which inevitably includes many background pixels. Fortunately, we can observe that the foreground pixels are meaningful and coherent compared to the uninformative background content within the bounding boxes.

Technically, we first compute the centroid for each source category in the feature space, which represents the mean feature vector of pixels belonging to the same object category within a source image. The formulation is defined as follows,
\begin{equation}
	c_s^k =\frac{1}{|I_s^k|}\sum\limits_{F_s^i\in{I_s^k}}F_s^i,\,k=\{1,2,\dots,K\}
\end{equation}
where $i$ is the pixel index, and $I_s^k$ denotes the set of pixels annotated with category $k$ in $F_s$. 
Then, we utilize the calculated $c_s^k$ to select foreground pixels in $I_s^k$ that satisfy the following two requirements:
\begin{itemize}
	\item For each source pixel $F_s^i$ in $I_s^k$, its similarity to the class centroid $c_s^k$ should larger than a certain value, \emph{i.e.,} $\cos(F_s^i,c_s^k)>\tau_1$, where $\cos(\cdot,\cdot)$ denotes the cosine similarity and $\tau_1$ is a threshold.
	\item To mitigate the influence of pixels that are located near the bounding box,
	the centerness~\cite{tian2019fcos} of pixel point $F_s^i$ should also larger than a certain value, \emph{i.e.,} 
	$\sqrt{\frac{\min(l,r)}{\max(l,r)}\times\frac{\min(t,b)}{\max(t,b)}}>\tau_2$, where $l$, $r$, $t$, and $b$ denote the normalized distances from the location of $F_s^i$ to the four sides of the bounding box, and $\tau_2$ is a threshold.
\end{itemize}
If the above requirements are all satisfied, $F_s^i$ is added into $\hat{I}_s^k$, 
where $\hat{I}_s^k$ is the selected foreground pixel set annotated with class $k$.
The overall source selected foreground pixel set is denoted by $\hat{I}_s=\{\hat{I}_s^k\}_{k=1}^K$.

Next, we aim to find pixels in $F_t$ that are more likely to be foreground pixels via \emph{cross-domain nearest neighbor search}.
We present the searching process below. For a source pixel $i$ in $\hat{I}_s^k$, we searched its nearest neighbor $j'$ in the target domain.
In turn, $i'$ is the nearest neighbor of target pixel $j'$ in the source domain.
If $i'$ also belongs to the category $k$, 
we will assign the target pixel $j'$ with pseudo-label $k$. 
Similarly, we traversal all pixels in $\hat{I}_s^k$ to find out pixel pairs between domains that are mutual nearest neighbors. 
In this way, we can obtain two set of selected pixels in both domains, \emph{i.e.,} $\hat{I}_s$ and $\hat{I}_t$. 
\subsubsection{Inter-domain Relational Reasoning}
\label{sec:3-1-2}
Given the searched foreground pixels in both source and target domains, we can easily project these visual features into node space, where each node corresponds to a foreground pixel in the feature map. Then, we aim to learn inter-domain visual dependencies in pixel level by introducing a bipartite graph convolutional module (BGCM) that can be inserted to detection model in a plug-and-play manner. To be specific, BGCM utilize a bipartite graph $\mathcal{G}_{\rm inter}^P = (\mathcal{V}_s^P, \mathcal{V}_t^P, \mathcal{E}^P)$ to indicate the relations among foreground nodes across domains and enhance their expressive power by aggregating information from its neighborhoods on the opposite domain. 
$\mathcal{V}_s^P$ and $\mathcal{V}_t^P$ are two group of bipartite graph nodes projected from $\hat{I}_s$ and $\hat{I}_t$ respectively.
Bipartite graph edges $\mathcal{E}^P$ represent the similarities between $\mathcal{V}_s^P$ and $\mathcal{V}_t^P$. 
To mitigate the impact of noisy background pixels and relations, we let the edge weights be learnable. 
Formally, for nodes $i$ and $j$ in $\mathcal{V}_s^P$ and $\mathcal{V}_t^P$, we have,    
\begin{equation}
	\mathcal{E}_{ij}^P = \sigma([F_s^i,F_t^j]\theta_e^P)
\end{equation}
where $\sigma$ stands for the sigmoid function, $F_s^i$ and $F_t^j$ are the features of nodes $i$ and $j$, and $\theta_e^P$ is the learnable parameter.

To perform graph convolution on the constructed bipartite graph $\mathcal{G}_P$, we augment its original form as follows,
\begin{equation}\label{eq:BGN}
	\hat{\mathcal{V}}^P=[\mathcal{V}_s^P, \mathcal{V}_t^P],
\end{equation} 
\begin{equation}\label{eq:BGE}
	\begin{aligned}
		\hat{\mathcal{E}}^P =
		\left(
		\begin{matrix}
			\mathbf{0} & \mathcal{E}^P  \\
			(\mathcal{E}^P)^T & \mathbf{0}\\
		\end{matrix}
		\right)
	\end{aligned}
\end{equation}
After that, the augmented bipartite graph $\hat{\mathcal{G}}_{\rm inter}^P=\{\hat{\mathcal{V}}^P,\hat{\mathcal{E}}^P\}$ can be learned by leveraging the modern Graph Convolutional Networks (GCN)~\cite{kipf2017semi}. In practice, we stack multiple graph convolution layers to form the BGCM. To be specific, the graph convolution is recursively conducted as: $\mX^{(l+1)} = {\rm ReLU}\left(\hat{\mA}\mX^{(l)}\mW^{(l)}\right)$, where $\mW^{l}$ is the parameter matrix, $\mX^{l}$ are the hidden features of the $l$-th layer (where $1\leq{l}\leq{L}$), and $\hat{\mA}$ is the adjacency matrix. 

To improve the discriminability of node features, we conduct node classification (\emph{i.e.} multi-class classification) based on the bipartite graph. 
Note that the selected source pixels have ground-truth labels, and we utilize the pixel-wise correspondence to assign pseudo-labels to the selected target pixels.
Formally, the last layer of pixel-level bipartite graph (BGCM) predicts the label using a classifier and can be formulated as follows, 
\begin{equation}\label{eq:cls}
	\hat{y}={\rm softmax}({FC}({BGCM}(x,\hat{\mathcal{G}}_{\rm inter}^P))),
\end{equation} 
where $\hat{y}$ denotes the predicted label, $FC$ denotes a fully-connected layer, and $x$ is the feature of source or target nodes. 
The node classification loss is represented by $\mathcal{L}_{\rm NC}$.
\subsubsection{Intra-domain Relational Reasoning}
\label{sec:3-1-3}
Although BGCM models the correlations between domains, the intra-domain topological structure is neglected.
We argue that relations within per domain are complementary to the inter-domain relations and should be explicitly taken into consideration.
To solve this issue, we introduce a graph attention module~(GAM), which allows nodes to attend over their neighborhoods features by specifying different weights to different neighbors, to perform intra-domain relational reasoning and capture the long-range dependencies between distant regions on each input image. 

Specifically, we directly construct the pixel-level intra-domain graph $\mathcal{G}_{\rm intra}^P=\{\mathcal{V}^P, \mathcal{E}_{\rm intra}^P\}$ on the source or target selected pixels ($\hat{I}_s$ or $\hat{I}_t$) to reduce the computational cost. Each node in $\mathcal{V}^P$ represents a selected pixel, and each edge in $\mathcal{E}_{\rm intra}^P$ encodes the relationships between nodes.
Then, we conduct self-attention on the nodes~\cite{velivckovic2018graph}, which computes the hidden representations that characterize the relationships between a source/target selected pixel and its neighborhoods within the domain. 
Given two nodes $i$ and $j$ in $\mathcal{G}_{\rm intra}^P$, their edge weights are defined as follows,
\begin{equation}\label{eq:intra_edge}
	\small
	\alpha_{ij}=\frac{\exp(\mathrm{LeakyReLU}(\va^T[\mW F^i \Vert \mW F^j]))}{\sum_{k \in \mathcal{N}_i}{\exp(\mathrm{LeakyReLU}(\va^T[\mW F^i \Vert \mW F^j]))}}
\end{equation} 
where $\mathcal{N}_i$ stands for the neighborhood of node $i$ in the graph, $\va$ is a learnable weight vector, $\mW$ is a weight matrix that is applied to every node, and $\Vert$ is the concatenation operation.

\textbf{Discussion.} Technically, graph self-attention is a type of \emph{local} attention, while self-attention in transformer~\cite{vaswani2017attention} is a type of \emph{global} attention. 
The constructed intra-domain relation graphs are sparse in the feature space. In that sense, local attention (based on the constructed graphs) will significantly reduce the potential of being biased towards those non-transferable nodes (pixels or regions), while global attention (based on the complete feature maps) is more likely to result in wrong feature aggregation and even negative transfer.

\subsection{Semantic-Level Relational Reasoning}
Instance-level (high-level) feature adaptation, which refers to aligning the ROI-based feature representations in the second stage of two-stage detectors, is a central step for learning semantic representations and mitigating high-level differences between domains, such as geometric variations (position, size, viewpoint, etc), the number of objects, and the type of objects.
In this regard, numerous instance-wise alignment modules has been proposed, such as vanilla adversarial alignment~\cite{chen2018domain}, weighted adversarial alignment~\cite{zhu2019adapting,He_2019_ICCV,chen2020harmonizing,xu2020exploring}, and prototype-guided alignment~\cite{xu2020cross,zheng2020cross}. 
However, these semantic alignment approaches suffer from two critical limitations. 
First, the alignment performance heavily relies on the quality of instance-level features, which are noisy and unpredictable in the target domain due to the lack of ground-truth annotations to distinguish foreground and background proposals during the ROI-pooling. 
Moreover, instance-level features are 
sensitive to the high-level variances, thereby hindering the deployment of direct alignment.  
Second, category alignment (one-vs-one) is insufficient to reason the foreground object dependencies since the graphical structure hidden in the semantic space (many-vs-many) lacks thorough investigation. Meanwhile, such high-level features are instance-informative yet domain-specific, requiring an elaborate calibration step to alleviate the feature noise.

To overcome the above challenges, we semantically explore the intra- and inter-domain object relations to enhance the reasoning ability of DAOD model by aggregating semantic and global context information via message-passing. In this section, we take Faster R-CNN as an exemplar to illustrate 
the technical details,
which can be readily applied to other modern detectors in experiments (cf. Section~\ref{sec:5-4}).

\subsubsection{Graph Nodes}
\label{sec:3-2-1}
Semantic-level graph nodes are introduced to represent the inherent characteristics of high-level visual concepts (object/instance/category). 
Regarding semantic-level graph, an intuitive idea is to utilize the proposals generated by RPN as graph nodes.
Unfortunately, these proposals may be insufficient to stand for an instance and cannot directly feed them into the relational reasoning procedure in view of the incomplete categorical information problems in the target domain, such as the deviation of bounding boxes, occlusions, and class ambiguities.
Moreover, due to the absence of target ground-truth bounding boxes, most existing DAOD works randomly sample target candidate proposals for ROI-pooling and fail to distinguish the positive and negative samples.
To solve this problem, we propose to use pseudo-labels for discriminating the target foreground instance-level features.

The workflow of Faster R-CNN can be divided into three components: backbone network, region proposal network (RPN), and ROI-wise classifier (RC). RPN generates and sends candidate object proposals to the head of RC for ROI-pooling based on the feature map extracted from the backbone network, then RC predicts the category labels and box coordinates of the pooled instance-level features. Both RPN and RC include a classification loss and a regression loss. 
Specifically, we utilize the detection results of RC to generate pseudo-labels for the target proposals. 
At the second stage, assume that we have $N_t$ target proposals filtered by RPN. The classification and regression outputs of RC are denoted by $F_{cls}(\boldsymbol{f}_p^i)$ and $B_p^i$, where $i=\{1,2,\dots,N_t\}$ and $\boldsymbol{f}_p^i$ represents the proposal feature. If $k'=\arg \max\limits_{k}\; F_{cls}(\boldsymbol{f}_p^i)$, this result will be added into class $k'$. Then, we sort the proposals within each class in descending order by their classification probabilities. 
A portion of high-confidence detection results in each class are selected as our pseudo labels, which are denoted by $\hat{B}_p^i$. 
In this way, we can ensure that the sampling process works in a class-balanced manner, implicitly highlighting the under-represented categories.
Finally, we utilize the pseudo-labeled bounding boxes to obtain corresponding proposal features from RPN, \emph{i.e.,}
$\boldsymbol{\hat{f}}_p^i={\rm \text{ROI-pooling}}(\boldsymbol{f}_g,\hat{B}_p^i)$.
These discriminative ROI-based instance-level features are the representations of semantic-level graph nodes, and the constructed graph can regarded be sparse representation of all region proposals.
Note that the target negative samples (backgrounds) are still randomly sampled from the remaining proposals.

\subsubsection{Inter-domain Relational Learning} 
\label{sec:3-2-2}
Next, we propose a semantic-level bipartite graph module (SBGM) to model semantic relations and constraints with respect to the foreground objects across domains.
The bipartite graph is defined as $\mathcal{G}_{\rm inter}^S=\{\mathcal{V}_s^S, \mathcal{V}_t^S, \mathcal{E}^S\}$. $\mathcal{V}_s^S = \{v_{s_i}\}_{i=1}^{N_p} \in \mathbb{R}^{N_p\times{d}}$ and $\mathcal{V}_t^S = \{v_{t_j}\}_{j=1}^{N_p} \in \mathbb{R}^{N_p\times{d}}$ are the source and target vertex sets, where $v_{s_i}$ and $v_{t_j}$ is the ROI-based instance-level features generated by RPN, $N_p$ is the number of proposals, $d$ is the node feature dimension, and $\mathcal{E}^S$ denotes the set of edges.

Given the graph nodes, SBGM aims to characterize the correspondence between $\mathcal{V}_s^S$ and $\mathcal{V}_t^S$.  
A natural idea is to traverse all possible pairs between $\mathcal{V}_s^S$ and $\mathcal{V}_t^S$ to compute their similarity, and node pairs with higher similarity should be assigned larger edge weights. 
Whilst we have enhanced the discriminability of foreground proposals, the instance-level features still contain positive and negative samples, and the distinction between known and unknown classes is yet to be explicitly constrained in light of the asymmetry of OSDA problem, making the target nodes be risky to aggregate biased or even wrong semantic information during the message-passing process. 

Consequently, to identify and isolate the background proposals, 
we design a Cross-Domain Similarity Regularization~(CDSR) strategy to generate reliable node pairs across domains. 
Our key idea is to adjust the cross-domain similarity measure to ensure that the nearest neighbor of a source node, in the target domain, is more likely to have as a nearest neighbor this particular source node, \textit{i.e.,} assign large edge weights to nodes from $\mathcal{V}_s^S$ and $\mathcal{V}_t^S$ that are mutual nearest neighbors. 
However, compared to the low-level feature space, we found that the nearest neighbors may be asymmetric in the high-level embedding space: 
$v_t$ (we omit the subscripts $i$ and $j$ for simplicity) being a $K$-NN of $v_s$ does not assure that $v_s$ is a $K$-NN of $v_t$, which also refer to the hubness problem~\cite{dinu2015improving,conneau2018word}. 
In high-level semantic space, 
some nodes are more likely to be the nearest neighbors of many other nodes (\emph{e.g.,} easy negatives), but some others may be not nearest neighbors of any node (\emph{e.g.,} hard positives).
For the bipartite graph $\mathcal{G}_{\rm inter}^S$, 
we denote the neighborhood of a source node $v_s$ as $\mathcal{N}_T(v_s)$. 
All $K$ elements of $\mathcal{N}_T(v_s)$ are nodes from $\mathcal{V}_t^S$. 
Likewise, the neighborhood associated with a target node $v_t$ is represented by $\mathcal{N}_S(v_t)$. 
All $K$ elements of $\mathcal{N}_S(v_t)$ are nodes from $\mathcal{V}_s^S$.
The average similarity of a source node $v_s$ to its target neighborhood is represented by,
\begin{equation}\label{eq:nn}
	r_T(v_s)=\frac{1}{K}\sum\limits_{v_t\in{\mathcal{N}_T(v_s)}}\cos(v_s,v_t),
\end{equation}
Similarly, the average similarity of a target node $v_t$ to its source neighborhood is represented by $r_S(v_t)$. 
Based on the computed $r_T(v_s)$ and $r_S(v_t)$, the cross-domain similarity measure ${\rm CDSR}(\cdot,\cdot)$ is formulated as follows,
\begin{equation}\label{eq:cdsr}
	{\rm CDSR}(v_s,v_t)={\rm sigmoid}(2\cos(v_s,v_t)-r_T(v_s)-r_S(v_t))
\end{equation}
Then, we can utilize Eq.~(\ref{eq:cdsr}) to calculate the adjacency matrix $\mA$, which associates each edge $(v_{s_i}, v_{t_j})$ with its element $\mA_{ij}$.
Finally, we use Eq.~(\ref{eq:BGN})-(\ref{eq:BGE}) to augment $\mathcal{G}_{\rm inter}^S$ as $\hat{\mathcal{G}}_{\rm inter}^S=\{\hat{\mathcal{V}}^S,\hat{\mathcal{E}}^S\}$ such that the modern graph convolution techniques can be conducted based on $\hat{\mathcal{G}}_{\rm inter}^S$.

To further enhance the semantic correlations among foreground objects between domains, we propose a Category-aware Domain Alignment~(CDA) loss term on top of $\hat{\mathcal{G}}_{\rm inter}^S$ to enable domain alignment on all foreground categories. 
Technically, we contrastively align the source and target prototypes to achieve domain alignment. Formally, we define the source and target class prototypes as follows,
\begin{equation}
	\begin{split}
		P_s^k = \frac{1}{|\hat{\mathcal{G}}_k^S|}\sum\limits_{x_s^i\in\hat{\mathcal{G}}_k^S}GCN_2(x_s^i,\hat{\mathcal{G}}_k^S) \\
		P_t^k = \frac{1}{|\hat{\mathcal{G}}_k^S|}\sum\limits_{x_t^i\in\hat{\mathcal{G}}_k^S}GCN_2(x_t^i,\hat{\mathcal{G}}_k^S)
	\end{split}
\end{equation} 
where $\hat{\mathcal{G}}_k^S$ stands for the nodes in $\mathcal{G}_{\rm inter}^S$ that belongs to category $k$ ($k\in\{1,2,\dots,K\}$). 
The target graph nodes can be clustered into $K$ classes by using the target pseudo-labels.
Formally, the formulation of CDA loss is defined as follows,
\begin{equation}\label{eq:CDA}
	\mathcal{L}_{\rm CDA}=\sum_k\left \| P_s^k - P_t^k \right \|_2 + \sum_{m\neq{n}}(max\{0, \xi-\left \| P_s^m, P_t^n \right \|_2\})
\end{equation}
where $\xi$ is the margin term and set as 1 in our experiments.

\subsubsection{Intra-domain Relational Learning}
\label{sec:3-2-3}
The semantic-level intra-domain graph are represented by $\mathcal{G}_{\rm intra}^S=\{\mathcal{V}^S, \mathcal{E}_{\rm intra}^S\}$.
The graph nodes are inherited from the SBGM. Each edge in $\mathcal{E}_{\rm intra}^S$ stands for the semantic correlations between $v_i$ and $v_j$.
To derive the intra-domain graph edge representations, we regularize the semantic relation from two aspects, i.e., geometric constraints and semantic similarities.
Specifically, intra-domain graph edge is represented by an adjacency matrix $\mA \in \mathbb{R}^{|\mathcal{V}^S|\times|\mathcal{V}^S|}$, which includes a spatial graph and a semantic graph.

For the spatial graph, we introduce $dist(B_p^i, B_p^j)$ and $\text{IoU}(B_p^i, B_p^j)$ to characterize the spatial interactions between nodes,
where $B_p^i$ and $B_p^j$ are two region proposals associated with two RoI-pooled instance-level features, $dist(B_p^i, B_p^j)$ denotes the distance between $B_p^i$ and $B_p^j$, and $\text{IoU}(B_p^i, B_p^j)$ measures their interactions in terms of Intersection over Union (IoU).
Formally, when $dist(B_p^i, B_p^j)<0.5$ or $\text{IoU}(B_p^i, B_p^j)>0.5$, $B_p^i$ and $B_p^j$ will be connected with a spatial graph edge (\emph{i.e.} $\mA_{ij}^{\rm spt} = 1$); otherwise, these two nodes will not be connected ($\mA_{ij}^{\rm spt} = 0$). By doing so, we can define the spatial adjacency matrix $\mA^{\rm spt}$.

For the semantic graph, we directly measure their cosine similarity to characterize the semantic interactions between nodes.
when $\cos(B_p^i, B_p^j)>0.5$, $B_p^i$ and $B_p^j$ will be connected with a semantic graph edge (\emph{i.e.} $\mA_{ij}^{\rm sec} = 1$); otherwise, these two nodes will not be connected ($\mA_{ij}^{\rm sec} = 0$). By doing so, we can define the spatial adjacency matrix $\mA^{\rm sec}$.

We integrate the spatial graph with the semantic graph to formulated our complete semantic-level intra-domain graph,
\begin{equation}\label{eq:intra_domain_graph_edge}
	\mA = \mA^{\rm spt} \circ \mA^{\rm sec}
\end{equation}
where $\circ$ stands for the element-wise dot. Following Eq.~(\ref{eq:intra_edge}), we conduct graph attention operation based on $\mathcal{G}_{\rm intra}^S$.

\subsection{Image-level Object-aware Reweighting} 
\label{sec:3-3}
In addition to the low-level (pixel-wise) and high-level (semantic-wise) features, we contend that the adaptation between global image-level features should be simultaneously considered now that image-level features, which contain rich information regarding the foreground objects, backgrounds, and scene layout, may be distinct across domains, \emph{i.e.,} the image-level features are not equally transferable. 
Conventional DAOD methods~\cite{chen2018domain,He_2019_ICCV,xu2020cross,zheng2020cross} commonly resort to fully align the whole image-level feature distributions of source and target data. 
Recent work of~\cite{saito2019strong,chen2020harmonizing} started to explore the adaptation difficulty of different images by assigning larger training weights to those easy-to-adapt images during the process of image-level adaptation. 

However, existing methods cannot deal with the object-level variations across domains based on the image-level features, such as the amount and category of foreground objects and the object co-occurrence, which is prone to result in severe feature misalignment or even negative transfer.
Moreover, as we usually utilize mini-batch SGD for training, the categorical information in each batch is insufficient to cover all objects.    
Thus, the object-level variations depend on the input image pairs and should be handled case-by-case.
Intuitively, we need to enlarge the training weights when the source and target samples contain similar content in the current batch; otherwise, the training weights will be reduced. Unfortunately, it is difficult to identify the category of objects in the target domain based on image-level features due to the absence of ground-truth annotations.

To remedy these issues, we propose a simple yet effective mechanism to characterize the degree of between-domain object-level variations from two aspects, \emph{i.e.,} the similarity of prototype representations and the number of identical object categories. 
The larger the object-level variation is, the smaller the training weight value is.
Given an input source and target image pair $x_s^i$ and $x_t^i$, we resort to the searched pixel-level correspondence in Section~\ref{sec:3-1-1} to compute these two terms. First, the weight regarding the similarity of prototype representations is formulated as follows, 
\begin{equation}\label{eq:w1}
	w_1=\sum_{N_k}\exp(-\lVert c_s^k - c_t^k \rVert^2) + 1
\end{equation}
where $N_k$ denotes the number of identical object categories between $x_s^i$ and $x_t^i$ based on the searched pixels, and $c_s^k$ and $c_t^k$ are the source and target prototype representations of category $k$.
Second, the weight with respect to the number of identical object categories is defined by,
\begin{equation}\label{eq:w2}
	w_2=\exp(\frac{N_k}{K})
\end{equation}
where $K$ is the number of classes and $N_k \leq K$.
Considering that the input image is treated as a whole in this step, no graph structure will be used here.
Thus, we decide to incorporate the computed training weights of images into a modern image-level adaptation objective. 
In this paper, we adopt the popular domain adversarial training loss~\cite{chen2018domain,saito2019strong,chen2020harmonizing}, which is formulated as follows,
\begin{equation}\label{eq:FAT}
	\begin{split}
		\mathcal{L}_{\text{IOR}}=&-\frac{1}{N_s}\sum\limits_{i=1}^{N_s}\frac{w_1+w_2}{2}\log(D_g(G(x_i^s))) \\
		&-\frac{1}{N_t}\sum\limits_{i=1}^{N_t}\frac{w_1+w_2}{2}\log(1-D_g(G(x_i^t)))
	\end{split}
\end{equation}
where $G$ is the backbone feature extractor and $D_g$ denotes image-wise domain discriminator. 

\subsection{Overall Objective}
The detection loss $\mathcal{L}_{\text{det}}$ contains a classification loss $\mathcal{L}_{\text{cls}}$ and a regression loss $\mathcal{L}_{\text{reg}}$. $\mathcal{L}_{\text{cls}}$ measures the classification accuracy of detector, and $\mathcal{L}_{\text{reg}}$ measures the degree of overlap between the ground-truth and predicted bounding boxes. 
Formally, the overall objective function for the proposed FGRR is defined as follows,
\begin{equation}\label{eq:all}
	\mathcal{L}_{\text{FGRR}}=\mathcal{L}_{\text{det}}+\lambda_1\mathcal{L}_{\text{NC}}+\lambda_2\mathcal{L}_{\rm CDA}+\lambda_3\mathcal{L}_{\text{IOR}}
\end{equation}
where $\lambda_1$, $\lambda_2$, and $\lambda_3$ are hyper-parameters for balancing the weights of different modules. 

\section{Theoretical Insight}
\label{sec:theory}
In this section, we investigate the connections between our approach and the theoretical upper bound of OSDA problem, making using of statistical learning theory of domain adaptation~\cite{fang2020open,ben2010theory,ben2010impossibility}. First, we provide the notations and problem setting of OSDA, and the definition of source and target risks. 
\begin{definition}\textbf{Open-Set Domain Adaptation (OSDA).}
	Suppose that we have a source domain $\mathcal{D}_s = \{(x_{s_i}, y_{s_i})\}_{i=1}^{n_s}$ of $n_s$ labeled samples and a target domain $\mathcal{D}_t = \{x_{t_j}\}_{j=1}^{n_t}$ of $n_t$ unlabeled samples. $\mathcal{D}_s$ and $\mathcal{D}_t$ are drawn from $P(\mathcal{X}_s, \mathcal{Y}_s)$ and $Q(\mathcal{X}_t, \mathcal{Y}_t)$, $P \neq Q$. The source and target label spaces share $K$ known classes and individually include a unknown class $u_s$ and $u_t$, which is different in both domains (i.e., $u_s \neq u_t$). 
	The goal of OSDA is to learn an optimal target classifier $h: \mathcal{X}_t\rightarrow \mathcal{Y}_t$.
\end{definition}

\begin{definition}\textbf{Source and Target Risks.}
	The source risk $R_s(h)$ and target risk $R_t(h)$ of $h$ w.r.t. $\mathcal{L}$ under source distribution $P$ and target distribution $Q$ are defined as
	\begin{equation}\notag
		\begin{split}
			R_s(h) &\triangleq \mathbb{E}_{(x, y)\sim P}\mathcal{L}(h(x), y) = \sum_{i=1}^{K+1}\pi_{i}^s R_{s,i}(h) \\
			R_t(h) &\triangleq \mathbb{E}_{(x, y)\sim Q}\mathcal{L}(h(x), y) = \sum_{j=1}^{K+1}\pi_{j}^t R_{t,j}(h)
		\end{split}
	\end{equation}
	where $\pi_i^s = P(y=i)$ and $\pi_j^t = Q(y=j)$ are class-prior probabilities of $P$ and $Q$. Then, we have 
	\begin{equation}\notag
		\begin{split}
			R_s(h) &= \sum_{i=1}^{K}\pi_{i}^s R_{s,i}(h)+\pi_{K+1}^s R_{s,K+1}(h)=R_s^*(h)+\Delta_{s} \\
			R_t(h) &= \sum_{j=1}^{K}\pi_{j}^t R_{t,j}(h)+\pi_{K+1}^t R_{t,K+1}(h)=R_t^*(h)+\Delta_{t}
		\end{split}
	\end{equation}
\end{definition}



Given the hypothesis space $\mathcal{H}$ with a condition that constant function $K + 1\in\mathcal{H}$, for $\forall h\in\mathcal{H}$, the expected error on target samples $R_t(h)$ is bounded as,
\begin{equation}\label{eq:OUDA}
	\small
	\begin{split}
		\frac{R_t(h)}{1-\pi_{K+1}^t} 
		&\leq R^*_s(h) + d_{\mathcal{H}\Delta\mathcal{H}}(P_{X|Y\leq K}, Q_{X|Y\leq K}) + \lambda \\
		&+\frac{\Delta_{t}}{1-\pi_{K+1}^t}			
	\end{split}
\end{equation}
where the shared error $\lambda =\min_{h\in\mathcal{H}}\frac{R_t^*(h)}{1-\pi_{K+1}^t} + R_s^*(h)$, $R^*_s(h)=\sum_{i=1}^{K}\pi_{i}^s R_{s,i}(h)$, and $\Delta_{t}=\pi_{K+1}^t R_{t,K+1}(h)$.
We show the derivation of Inequality~(\ref{eq:OUDA}) in the supplementary material. 
From the inequality, we can see that the target error is bounded by the following four terms: (1) expected error on the known classes of source domain $R^*_s(h)$; (2) domain discrepancy $d_{\mathcal{H}\Delta\mathcal{H}}(P_{X|Y\leq C}, Q_{X|Y\leq C})$; (3) shared error $\lambda$ of the ideal joint hypothesis $h^*$; (4) target open set risk $\Delta_{t}$.  
\begin{remark}
	$R^*_s(h)$, which can be optimized by using the source ground-truth annotations, is expected to be small. $d_{\mathcal{H}\Delta\mathcal{H}}(P_{X|Y\leq C}, Q_{X|Y\leq C})$, which measures the discrepancy between source and target data distributions, can be optimized by using any modern domain alignment approaches. 
	$\lambda$, which reflects the category-level conditional shift, is considered to be sufficiently small since our method explicitly model the semantic correlations.
	The target open set risk $\Delta_{t}$ is prone to be large when an approach does not make a distinction between the target foregrounds (known classes) and backgrounds (unknown class). 
	By contrast, our approach optimize this term by discriminating the foreground pixels and regions and performing foreground-aware relational reasoning.
	To summarize, the proposed FGRR is capable of effectively optimizing the upper bound of expected target error by simultaneously minimizing the above four terms. 	
\end{remark}

\section{Experiments}
\label{sec:experiments}
In this section, we conduct experiments on four domain shifts including seven natural image datasets and two medical image datasets, 
\textit{i.e.,} real-to-artistic (Pascal VOC~$\rightarrow$~Clipart, Pascal VOC~$\rightarrow$~Watercolor, and Pascal VOC~$\rightarrow$~Comic), normal-to-foggy (Cityscapes~$\rightarrow$~Foggy-Cityscapes), synthetic-to-real (Sim10k~$\rightarrow$~Cityscapes), and cross-site mammogram mass detection (Public~$\rightarrow$~In-house).  

\subsection{Dataset}
\textbf{Real-to-Artistic.} In this case, we combine the Pascal VOC2007-trainval and VOC2012-trainval datasets as the source domain, and use Clipart1k, Watercolor2k, and Comic2k as the target domains respectively. The Pascal VOC~\cite{everingham2010pascal} is a real-world image dataset, which contains 16,551 images with 20 object classes. 
Clipart1k, Watercolor2k, and Comic2k, which are collected from a website called Behance and annotated by Inoue~\textit{et al.} for cross-domain object detection tasks, consist of 1,000, 2,000, and 2,000 images respectively. 
Clipart1k has the same 20 object categories as Pascal VOC, and Watercolor2k and Comic2k share 6 identical object classes with the Clipart1k dataset, \emph{i.e.,} bicycle, bird, cat, car, dog, and person. 
For Pascal VOC~$\rightarrow$~Clipart, we use all images of Clipart1k as the target domain for both training and testing by following mainstream DAOD works~\cite{saito2019strong,chen2020harmonizing}. 
For Pascal VOC~$\rightarrow$~Watercolor and Pascal VOC~$\rightarrow$~Comic, we leverage the train set (1K images) for training and the test set (1K images) is held out for evaluation. 

\noindent
\textbf{Normal-to-Foggy.} The images in Cityscapes~\cite{cordts2016cityscapes} are captured from the street scenes of different cities in normal weather conditions via a car-mounted video camera.  
Since the dataset has dense pixel-level labels, we use the rectangle of instance mask to derive ground-truth bounding boxes for our DAOD tasks. Foggy-Cityscapes~\cite{cordts2016cityscapes} are rendered from Cityscapes by using the depth maps to simulate the foggy scenes, and the bounding box annotations are naturally inherited from the original Cityscapes dataset. 
Both datasets own 2,975 images in the training set and 500 images in the validation set. 
Note that although images in Cityscapes and Foggy-Cityscapes have a one-to-one correspondence, we do not use this information when training DAOD models.

\noindent
\textbf{Synthetic-to-Real.} Sim10K~\cite{johnson2017driving} is a driving scene dataset that was produced based on the computer game Grand Theft Auto V (GTA V). It consists of 10,000 synthetic images with 58,071 bounding boxes of the car. We use all images in Sim10K as the source domain. In addition, the Cityscapes dataset, which contains real road scene images, is used as the target domain. Sim10K and Cityscapes share one identical object class, \textit{i.e.,} car. 

\noindent
\textbf{Cross-Site Mammogram Mass Detection.} We further evaluate our model on two mammograms datasets with labeled masses: a public dataset (INbreast~\cite{moreira2012inbreast}) and an in-house dataset. INbreast, which includes 107 low-quality mammography cases with 112 labeled masses, is one of the most broadly-used mammogram mass detection datasets. The in-house dataset, which includes 297 high-quality cases with 448 labeled masses, is collected from a local hospital.
The annotations are labeled by two radiologists with strong expertise.
We use these two datasets to build a DAOD task: Public~$\rightarrow$~In-house.

\subsection{Implementation Details}
We follow the same setting in mainstream DAOD methods~\cite{chen2018domain,saito2019strong,chen2020harmonizing} that utilize Faster-RCNN~\cite{ren2015faster} with VGG-16~\cite{simonyan2014very} or ResNet-101~\cite{he2016deep} architectures as the detection model. We fine-tune VGG-16 and ResNet-101 pretrained on ImageNet.
In all experiments, the shorter side of the image is resized to 600. 
The batch size is selected as 2 (one image per domain) to fit the GPU memory.
In the testing phase, we report mean average precision (mAP) with a IoU threshold of 0.5 on the target domain to evaluate the adaptation performance. 
We train the proposed model using stochastic gradient descent (SGD) optimizer with an initial learning rate of 0.001 and momentum 0.9.
The learning rate is decreased to 0.0001 after 5 epochs.
For the hyper-parameters $\lambda_1$, $\lambda_2$, and $\lambda_3$ in Eq.~(\ref{eq:all}), We set $\lambda_1=\lambda_2=0.1$ and $\lambda_3=1$ in all experiments considering that we do not have access to the target labels in both model training and selection phase.	
The experiments are implemented based on PyTorch deep learning framework.

\begin{center}
	\begin{table*}[htb]
		\caption{Results on adaptation from PASCAL VOC to Clipart Dataset (\%). \textbf{mAP} is reported on the target domain. \textbf{mAP*} denotes the result of Source Only model for each method, and \textbf{Gain} stands for its the improvement after adaptation.}\label{table1}
		\vspace{-0.2cm}
		\centering
		\footnotesize
		\scalebox{0.9}{
			\setlength\tabcolsep{2pt}
			\begin{tabular}{c|c|cccccccccccccccccccc|cc|a}
				\toprule
				Methods & backbone & \rotatebox[origin=c]{90}{aero} & \rotatebox[origin=c]{90}{bcycle} & \rotatebox[origin=c]{90}{bird} & \rotatebox[origin=c]{90}{boat} & \rotatebox[origin=c]{90}{bottle} & \rotatebox[origin=c]{90}{bus} & \rotatebox[origin=c]{90}{car} & \rotatebox[origin=c]{90}{cat} & \rotatebox[origin=c]{90}{chair} & \rotatebox[origin=c]{90}{cow} & \rotatebox[origin=c]{90}{table} & \rotatebox[origin=c]{90}{dog} & \rotatebox[origin=c]{90}{hrs} & \rotatebox[origin=c]{90}{bike} & \rotatebox[origin=c]{90}{prsn} & \rotatebox[origin=c]{90}{plnt} & \rotatebox[origin=c]{90}{sheep} & \rotatebox[origin=c]{90}{sofa} & \rotatebox[origin=c]{90}{train} & \rotatebox[origin=c]{90}{tv} & mAP & mAP* & Gain \\
				\hline
				SA-DA-Faster~\cite{chen2021scale} & ResNet-50-FPN & 29.4 & 56.8 & 30.6 & 34.0 & 49.5 & 50.5 & 47.7 & 18.7 & 48.5 & 64.4 & 20.3 & 29.0 & 42.3 & 84.1 & 73.4 & 37.4 & 20.5 & 39.8 & 41.2 & 48.0 & 43.3 & 28.8 & 14.5 \\
				\hline
				DA-Faster~\cite{chen2018domain} & ResNet-101 & 15.0 & 34.6 & 12.4 & 11.9 & 19.8 & 21.1 & 23.2 & 3.1 & 22.1 & 26.3 & 10.6 & 10.0 & 19.6 & 39.4 & 34.6 & 29.3 & 1.0 & 17.1 & 19.7 & 24.8 & 19.8 & 27.8 & -8.0 \\
				MAF~\cite{He_2019_ICCV} & ResNet-101 & 38.1 & 61.1 & 25.8 & 43.9 & 40.3 & 41.6 & 40.3 & 9.2 & 37.1 & 48.4 & 24.2 & 13.4 & 36.4 & 52.7 & 57.0 & \textbf{\color{blush}52.5} & 18.2 & 24.3 & 32.9 & 39.3 & 36.8 & 27.8 & 9.0 \\
				SWDA~\cite{saito2019strong} & ResNet-101 & 26.2 & 48.5 & 32.6 & 33.7 & 38.5 & 54.3 & 37.1 & 18.6 & 34.8 & 58.3 & 17.0 & 12.5 & 33.8 & 65.5 & 61.6 & 52.0 & 9.3 & 24.9 & 54.1 & 49.1 & 38.1 & 27.8 & 10.3 \\
				ICR-CCR~\cite{xu2020exploring} & ResNet-101 & 28.7 & 55.3 & 31.8 & 26.0 & 40.1 & 63.6 & 36.6 & 9.4 & 38.7 & 49.3 & 17.6 & 14.1 & 33.3 & 74.3 & 61.3 & 46.3 & 22.3 & 24.3 & 49.1 & 44.3 & 38.3 & 27.0 & 11.3 \\
				HTCN~\cite{chen2020harmonizing} & ResNet-101 & 33.6 & 58.9 & 34.0 & 23.4 & \textbf{\color{blush}45.6} & 57.0 & 39.8 & 12.0 & 39.7 & 51.3 & 21.1 & 20.1 & 39.1 & 72.8 & 63.0 & 43.1 & 19.3 & 30.1 & 50.2 & \textbf{\color{blush}51.8} & 40.3 & 27.8 & 12.5 \\
				CDTD~\cite{shen2021cdtd} & ResNet-101 & \textbf{\color{blush}44.7} & 50.0 & 33.6 & 27.4 & 42.2 & 55.6 & 38.3 & 19.2 & 37.9 & 69.0 & 30.1 &  26.3 & 34.4 & 67.3 & 61.0 & 47.9 & 21.4 & 26.3 & 50.1 & 47.3 & 41.5 & 27.8 & 13.7 \\
				ATF~\cite{he2020domain} & ResNet-101 & 41.9 & \textbf{\color{blush}67.0} & 27.4 & \textbf{\color{blush}36.4} & 41.0 & 48.5 & 42.0 & 13.1 & 39.2 & \textbf{\color{blush}75.1} & 33.4 & 7.9 & \textbf{\color{blush}41.2} & 56.2 & 61.4 & 50.6 & 42.0 & 25.0 & 53.1 & 39.1 & 42.1 & 27.8 & 14.3 \\
				PD~\cite{wu2021instance} & ResNet-101 & 41.5 & 52.7 & 34.5 & 28.1 & 43.7 & 58.5 & 41.8 & 15.3 & 40.1 & 54.4 & 26.7 & \textbf{\color{blush}28.5} & 37.7 & 75.4 & 63.7 & 48.7 & 16.5 & 30.8 & 54.5 & 48.7 & 42.1 & 27.8 & 14.3 \\
				\hline
				DBGL~\cite{chen2021dual} & ResNet-101 & 28.5 & 52.3 & 34.3 & 32.8 & 38.6 & \textbf{\color{blush}66.4} & 38.2 & \textbf{\color{blush}25.3} & 39.9 & 47.4 & 23.9 & 17.9 & 38.9 & 78.3 & 61.2 & 51.7 & \textbf{\color{blush}26.2} & 28.9 & 56.8 & 44.5 & 41.6 & 27.8 & 13.8 \\ 
				FGRR (Ours) & ResNet-101 & 30.8 & 52.1 & \textbf{\color{blush}35.1} & 32.4 & 42.2 & 62.8 & \textbf{\color{blush}42.6} & 21.4 & \textbf{\color{blush}42.8} & 58.6 & \textbf{\color{blush}33.5} & 20.8 & 37.2 & \textbf{\color{blush}81.4} & \textbf{\color{blush}66.2} & 50.3 & 21.5 & 29.3 & \textbf{\color{blush}58.2} & 47.0 & \textbf{\color{blush}43.3} & 27.8 & \textbf{\color{blush}15.5} \\
				\bottomrule
		\end{tabular}}
	\end{table*}
\end{center}

\begin{center}
	\begin{table*}[thb]
		\caption{Results on adaptation from Cityscapes to Foggy-Cityscapes (\%).}\label{table2}
		\centering
				\begin{tabular}{c|c|cccccccc|cc|a}
					\toprule
					Methods & Backbone & Person & Rider & Car & Truck & Bus & Train & Motorbike & Bicycle & mAP & mAP* & Gain \\
					\hline
					\hline
					MTOR~\cite{cai2019exploring} & ResNet-50 & 30.6 & 41.4 & 44.0 & 21.9 & 38.6 & 40.6 & 28.3 & 35.6 & 35.1 & 26.9 & 8.2 \\
					GPA~\cite{xu2020cross} & ResNet-50 & 32.9 & 46.7 & 54.1 & 24.7 & 45.7 & 41.1 & 32.4 & 38.7 & 39.5 & 26.9 & 12.6 \\
					PD~\cite{wu2021instance} & ResNet-101 & 32.8 & 44.4 & 49.6 & 33.0 & 46.1 & 38.0 & 29.9 & 35.3 & 38.6 & 25.6 & 13.0 \\
					AFAN~\cite{wang2021afan} & ResNet-50-FPN & 42.5 & 44.6 & 57.0 & 26.4 & 48.0 & 28.3 & 33.2 & 37.1 & 39.6 & 26.1 & 13.5 \\
					SA-DA-Faster~\cite{chen2021scale} & ResNet-50-FPN & 50.3 & 45.4 & 62.1 & 32.4 & 48.5 & 52.6 & 31.5 & 29.5 & 44.0 & 30.3 & 13.7 \\
					\hline
					SCDA~\cite{zhu2019adapting} & VGG-16 & 33.5 & 38.0 & 48.5 & 26.5 & 39.0 & 23.3 & 28.0 & 33.6 & 33.8 & 26.2 & 7.6 \\
					DA-Faster~\cite{chen2018domain} & VGG-16 & 25.0 & 31.0 & 40.5 & 22.1 & 35.3 & 20.2 & 20.0 & 27.1 & 27.6 & 18.8 & 8.8 \\
					CT~\cite{zhao2020collaborative} & VGG-16 & 32.7 & 44.4 & 50.1 & 21.7 & 45.6 & 25.4 & 30.1 & 36.8 & 35.9 & 26.2 & 9.7 \\
					CDN~\cite{su2020adapting} & VGG-16 & 35.8 & 45.7 & 50.9 & 30.1 & 42.5 & 29.8 & 30.8 & 36.5 & 36.6 & 26.1 & 10.5 \\
					PD~\cite{wu2021instance} & VGG-16 & 33.1 & 43.4 & 49.6 & 22.0 & 45.8 & 32.0 & 29.6 & 37.1 & 36.6 & 22.8 & 13.8 \\
					SWDA~\cite{saito2019strong} & VGG-16 & 29.9 & 42.3 & 43.5 & 24.5 & 36.2 & 32.6 & 30.0 & 35.3 & 34.3 & 20.3 & 14.0 \\
					PAL~\cite{sindagi2020prior} & VGG-16 & 36.4 & 47.3 & 51.7 & 22.8 & 47.6 & 34.1 & \textbf{\color{blush}36.0} & 38.7 & 39.3 & 24.4 & 14.9 \\
					MAF~\cite{He_2019_ICCV} & VGG-16 & 28.2 & 39.5 & 43.9 & 23.8 & 39.9 & 33.3 & 29.2 & 33.9 & 34.0 & 18.8 & 15.2 \\
					ICR-CCR~\cite{xu2020exploring} & VGG-16 & \textbf{\color{blush}45.1} & 34.6 & 49.2 & 30.3 & 32.9 & 43.8 & 36.4 & 27.2 & 37.4 & 22.0 & 15.4 \\
					DD-MRL~\cite{kim2019diversify} & VGG-16 & 30.8 & 40.5 & 44.3 & 27.2 & 38.4 & 34.5 & 28.4 & 32.2 & 34.6 & 17.9 & 16.7 \\
					VDD~\cite{wu2021vector} & VGG-16 & 33.4 & 44.0 & 51.7 & 33.9 & \textbf{\color{blush}52.0} & 34.7 & 34.2 & 36.8 & 40.0 & 22.8 & 17.2 \\
					MeGA-CDA~\cite{vs2021mega} & VGG-16 & 37.7 & \textbf{\color{blush}49.0} & \textbf{\color{blush}52.4} & 25.4 & 49.2 & 46.9 & 34.5 & \textbf{\color{blush}39.0} & \textbf{\color{blush}41.8} & 24.4 & 17.4 \\
					CDTD~\cite{shen2021cdtd} & VGG-16 & 31.6 & 44.0 & 44.8 & 30.4 & 41.8 & 40.7 & 33.6 & 36.2 & 37.9 & 20.3 & 17.6 \\
					C2F~\cite{zheng2020cross} & VGG-16 & 43.2 & 37.4 & 52.1 & \textbf{\color{blush}34.7} & 34.0 & \textbf{\color{blush}46.9} & 29.9 & 30.8 & 38.6 & 20.8 & 17.8 \\
					ATF~\cite{he2020domain} & VGG-16 & 34.6 & 47.0 & 50.0 & 23.7 & 43.3 & 38.7 & 33.4 & 38.8 & 38.7 & 20.3 & 18.4 \\
					HTCN~\cite{chen2020harmonizing} & VGG-16 & 33.2 & 47.5 & 47.9 & 31.6 & 47.4 & 40.9 & 32.3 & 37.1 & 39.8 & 20.3 & 19.5 \\
					\hline
					\hline
					DBGL~\cite{chen2021dual} & VGG-16 & 33.5 & 46.4 & 49.7 & 28.2 & 45.9 & 39.7 & 34.8 & 38.3 & 39.6 & 20.3 & 19.3 \\
					FGRR (Ours) & VGG-16 & 34.4 & 47.6 & 51.3 & 30.0 & 46.8 & 42.3 & 35.1 & 38.9 & 40.8 & 20.3 & \textbf{\color{blush}20.5} \\
					\bottomrule
				\end{tabular}
			\end{table*}
		\end{center}
		
		\begin{center}
			\begin{table*}[htb]
				\caption{Results on Pascal VOC $\rightarrow$ Watercolor2k (\%). mAP is reported on the Watercolor2k test set.}\label{table3}
				\centering
				\small
				\begin{tabular}{c|c|cccccc|cc|a}
					\toprule
					Methods & Backbone & bike & bird & car & cat & dog & person & mAP & mAP* & Gain \\
					\hline
					\hline
					AFAN~\cite{wang2021afan} & ResNet-50-FPN & 87.0 & 46.4 & 47.3 & 33.1 & 30.0 & 60.1 & 50.6 & 43.1 & 7.5 \\
					\hline
					BDC-Faster & ResNet-101 & 68.6 & 48.3 & 47.2 & 26.5 & 21.7 & 60.5 & 45.5 & 44.6 & 0.9 \\
					DA-Faster~\cite{chen2018domain} & ResNet-101 & 75.2 &40.6 &48.0 &31.5 &20.6 &60.0 &46.0 & 44.6 & 1.4 \\
					MAF~\cite{He_2019_ICCV} & ResNet-101 & 73.4 & 55.7 & 46.4 & 36.8 & 28.9 & 60.8 & 50.3 & 44.6  & 5.7 \\
					SWDA~\cite{saito2019strong} & ResNet-101 & 82.3 & 55.9 &46.5 & 32.7 & 35.5 & 66.7 & 53.3 & 44.6 & 8.7 \\
					ATF~\cite{he2020domain} & ResNet-101 & 78.8 & \textbf{\color{blush}59.9} & 47.9 & \textbf{\color{blush}41.0} & 34.8 & 66.9 & 54.9 & 44.6 & 10.3 \\
					CDTD~\cite{shen2021cdtd} & ResNet-101 & 82.2 & 55.1 & \textbf{\color{blush}51.8} & 39.6 & 38.4 & 64.0 & 55.2 & 44.6 & 10.6 \\
					\hline
					\hline
					DBGL~\cite{chen2021dual} & ResNet-101 & 83.1 & 49.3 & 50.6 & 39.8 & 38.7 & 61.3 & 53.8 & 44.6 & 9.2 \\
					FGRR (Ours) & ResNet-101 & \textbf{\color{blush}86.1} & 54.8 & 48.9 & 36.6 & \textbf{\color{blush}40.4} & \textbf{\color{blush}67.5} & \textbf{\color{blush}55.7} & 44.6 & \textbf{\color{blush}11.1} \\
					\bottomrule
				\end{tabular}
			\end{table*}
		\end{center}
		
		\begin{center}
			\begin{table}[!t]
				\caption{Results on Pascal VOC $\rightarrow$ Comic2k (\%). The backbone network is ResNet-101.}\label{table4}
				\centering
				\setlength\tabcolsep{3pt}
				\begin{tabular}{c|cccccc|cc|a}
					\toprule
					Methods & \rotatebox[origin=c]{90}{bike} & \rotatebox[origin=c]{90}{bird} & \rotatebox[origin=c]{90}{car} & \rotatebox[origin=c]{90}{cat} & \rotatebox[origin=c]{90}{dog} & \rotatebox[origin=c]{90}{person} & mAP & mAP* & Gains \\
					\hline
					SWDA~\cite{saito2019strong} & 36.0 & 18.3 & 29.3 & 9.3 & 22.9 & 48.4 & 27.4 & 24.4 & 3.0 \\
					\hline
					DBGL~\cite{chen2021dual} & 35.6 & 20.3 & \textbf{\color{blush}33.9} & 16.4 & 26.6 & 45.3 & 29.7 & 24.4 & 5.3 \\
					FGRR (Ours) & \textbf{\color{blush}42.2} & \textbf{\color{blush}21.1} & 30.2 & \textbf{\color{blush}21.9} & \textbf{\color{blush}30.0} & \textbf{\color{blush}50.5} & \textbf{\color{blush}32.7} & 24.4 & \textbf{\color{blush}8.3} \\
					\bottomrule
				\end{tabular}
			\end{table}
		\end{center}
		
		\begin{center}
			\begin{table}[htb]
				\caption{Results on Sim10K~$\rightarrow$~Cityscapes (\%). \textbf{AP*} denotes the results of Source Only model for each method.}\label{table5}
				\centering
				\setlength\tabcolsep{5pt}
				\scalebox{0.95}{
					\begin{tabular}{c|c|cc|a}
						\toprule
						Methods & Backbone & AP on \emph{car} & AP* on \emph{car} & Gain \\
						\hline
						\hline
						MTOR~\cite{cai2019exploring} & ResNet-50 & 46.6 & 39.4 & 7.2 \\
						AFAN~\cite{wang2021afan} & ResNet-50-FPN & 45.5 & 32.9 & 12.6 \\
						SA-DA-Faster~\cite{chen2021scale} & ResNet-50-FPN & 55.8 & 36.7 & \textbf{\color{blush}19.1} \\ 
						\hline  
						DA-Faster~\cite{chen2018domain} & VGG-16 & 38.9 & 34.6 & 4.3 \\
						SWDA~\cite{saito2019strong} & VGG-16 & 40.1 & 34.6 & 5.5 \\
						HTCN~\cite{chen2020harmonizing} & VGG-16 & 42.5 & 34.6 & 7.9 \\
						CDTD~\cite{shen2021cdtd} & VGG-16 & 42.6 & 34.6 & 8.0 \\
						ATF~\cite{he2020domain} & VGG-16 & 42.8 & 34.6 & 8.2 \\
						UMT~\cite{deng2021unbiased} & VGG-16 & 43.1 & 34.3 & 8.8 \\
						MeGA-CDA~\cite{vs2021mega} & VGG-16 & \textbf{\color{blush}44.8} & 34.3 & 10.5 \\
						MAF~\cite{He_2019_ICCV} & VGG-16 & 41.1 & 30.1 & 11.0 \\
						Noise Labeling~\cite{khodabandeh2019robust} & VGG-16 & 42.6 & 31.1 & 11.5 \\
						\hline
						\hline
						DBGL~\cite{chen2021dual} & VGG-16 & 42.7 & 34.6 & 8.1 \\
						Ours & VGG-16 & 44.5 & 34.6 & 9.9 \\
						\bottomrule
				\end{tabular}}
			\end{table}
		\end{center}
		
		\begin{center}
			\begin{table}[htb]
				\caption{Results on Public~$\rightarrow$~In-house (\%). The backbone network is ResNet-101.}\label{table6}
				\centering
				\setlength\tabcolsep{5pt}
					\begin{tabular}{c|cc|a}
						\toprule
						Methods & AP on \emph{mass} & AP* on \emph{mass} & Gain \\
						\hline
						DA-Faster~\cite{chen2018domain} & 20.2 & 12.0 & 8.2 \\
						SWDA~\cite{saito2019strong} & 27.9 & 12.0 & 15.9 \\
						MAF~\cite{He_2019_ICCV} & 33.1 & 12.0 & 21.1 \\
						HTCN~\cite{chen2020harmonizing} & 36.0 & 12.0 & 24.0 \\
						ATF~\cite{he2020domain} & 37.5 & 12.0 & 25.5 \\
						\hline
						DBGL~\cite{chen2021dual} & 38.4 & 12.0 & 26.4 \\
						FGRR (Ours) & \textbf{\color{blush}43.9} & 12.0 & \textbf{\color{blush}31.9} \\
						\bottomrule
					\end{tabular}
				\end{table}
			\end{center}

			\vspace{-4cm}
			\subsection{Comparisons with State-of-the-Arts}
			\subsubsection{Comparison Methods}
			We compare the proposed FGRR with state-of-the-art DAOD methods. 
			(1) \emph{Domain alignment:} 
			Domain adaptive Faster-RCNN (\textbf{DA-Faster})~\cite{chen2018domain}, 
			Strong-Weak Distribution Alignment~(\textbf{SWDA})~\cite{saito2019strong},  
			Domain Diversification and Multi-domain-invariant Representation Learning (\textbf{DD-MRL})~\cite{kim2019diversify},
			Collaborative Training (\textbf{CT})~\cite{zhao2020collaborative}, 
			Augmented Feature Alignment Network~\textbf{(AFAN)}~\cite{wang2021afan}, 
			\textbf{CDTD}~\cite{shen2021cdtd}, 
			and Scale-Aware DA-Faster~\textbf{(SA-DA-Faster)}~\cite{chen2021scale}.
			(2) \emph{Local alignment:}
			Mean Teacher with Object Relations (\textbf{MTOR})~\cite{cai2019exploring}, 
			Selective Cross-Domain Alignment~(\textbf{SCDA})~\cite{zhu2019adapting}, 
			Multi-Adversarial Faster-RCNN (\textbf{MAF})~\cite{He_2019_ICCV},
			Conditional Domain Normalization (\textbf{CDN})~\cite{su2020adapting}, 
			Progressive Disentanglement (\textbf{PD})~\cite{wu2021instance}, 
			Image-level Categorical Regularization and Categorical Consistency Regularization (\textbf{ICR-CCR})~\cite{xu2020exploring}, Coarse-to-Fine (\textbf{C2F})~\cite{zheng2020cross}, 
			Asymmetric Tri-way Faster-RCNN (\textbf{ATF})~\cite{he2020domain}, 
			Graph-induced Prototype Alignment (\textbf{GPA})~\cite{xu2020cross}, 
			Hierarchical Transferability Calibration Network (\textbf{HTCN})~\cite{chen2020harmonizing}, 
			Prior-Adversarial Loss~\textbf{(PAL)}~\cite{sindagi2020prior}, 
			Memory Guided Attention for Category-Aware Domain Adaptation~\textbf{(MeGA-CDA)}~\cite{vs2021mega},  
			Vector-Decomposed Disentanglement (\textbf{VDD})~\cite{wu2021vector},
			and Dual Bipartite Graph Learning (\textbf{DBGL})~\cite{chen2021dual}. 
			
			For all the aforementioned methods, we cite the quantitative results from their original papers. \textbf{Source Only} stands for the model that is trained only using source images and directly evaluated on the target domain without any adaptation. 
			To facilitate a fair comparison, we utilize the gain of performance compared to the Source Only model as the main evaluation metric by following~\cite{zheng2020cross}. 
			
			\subsubsection{Results and Discussion}
			\textbf{Real-to-Artistic.} Table~\ref{table1}, Table~\ref{table3}, and Table~\ref{table4} report the adaptation results of Pascal VOC~$\rightarrow$~Clipart, Pascal VOC~$\rightarrow$~Watercolor, and Pascal VOC~$\rightarrow$~Comic. The proposed FGRR significantly outperforms all comparison methods on different real-to-artistic adaptation tasks. 
			In particular, compared to the preliminary version (DBGL), the proposed FGRR improves over its results by +2.2\% on average (41.6\% $\rightarrow$ 43.3\%, 53.8\% $\rightarrow$ 55.7\%, and 29.7\% $\rightarrow$ 32.7\%), which clearly demonstrates the significance of our improvements. In contrast to adversarial-based domain alignment (\emph{e.g.,} DA-Faster and SWDA) and local alignment approaches (\emph{e.g.,} MAF, HTCN, MeGA-CDA, and VDD), FGRR additionally explore the topological interactions on both intra- and inter-domains to enable fine-grained knowledge transfer. 
			Our better performance over them can explicitly testify the effectiveness of these two components. 
			ATF resorts to the tri-training strategy to solve the labeling unfairness problem between source and target domains, and thus is complementary to our FGRR. 
			That is to say, our work focuses on adapting internal features without requiring model ensemble (\emph{e.g.,} mean-teacher~\cite{cai2019exploring} and tri-training~\cite{he2020domain}).
			The better performance of FGRR further verifies the importance of endowing DAOD model with the relational reasoning ability. 
			In addition, since the proposed model explicitly consider the imbalance problem in DAOD, the per-class adaptation performance is more balanced.
			For instance, in Table~\ref{table4}, the performances of ``bird'', ``car'' ``cat'', and ``dog'' are balanced, and their average performance is substantially improved at the same time.
			
			\textbf{Normal-to-Foggy.} 
			Table~\ref{table2} displays the results on adaptation from Cityscapes to Foggy-Cityscapes.
			This scenario is the most widely used DAOD datasets, and contains distinct object co-occurrence, backgrounds, and weather. The source and target images are originally the same one, which makes this task more suitable for explicit local feature alignment approaches, such as HTCN~\cite{chen2020harmonizing} and MeGA-CDA~\cite{vs2021mega}.  
			As can be seen, the proposed FGRR achieves a remarkable increase of +20.5\% over the Source Only model on adaptation between these two similar domains, 
			verifying the robustness and effectiveness of FGRR on the different DAOD scenario.
			The results also reveal several interesting observations. 
			\begin{enumerate}
				\item The series of holistic feature alignment approaches (such as DA-Faster, SWDA, and DD-MRL) demonstrate inferior performance since they do not consider the local nature of object detection problem. In other words, the objects of interest only take a small portion of the whole image, and thus globally aligning features between domains is prone to result in negative transfer. 
				
				\item Local alignment approaches (such as SCDA, C2F, MeGA-CDA, HTCN, and VDD) change the focus of the adaptation from holistic to local by virtue of elaborate local alignment methods regarding the foreground objects. In this way, they substantially improve the adaptation performance compared to holistic alignment approaches. 
				
				\item MTOR explores the object relation within and between domains in the context of a mean-teacher framework. However, they focus on optimizing consistency regularization without delving into the interactions and dependencies of different foreground objects. By contrast, the proposed FGRR utilizes the bipartite graph to model the cross-domain relations where two set of features are aggregated over the spatial and semantic spaces. 
				Noting that FGRR significantly outperforms the result of MTOR by 5.7\% (from 35.1\% to 40.8\%).
				
				\item C2F and GPA aim to achieve the cross-domain semantic consistency based on prototype alignment. 
				In particular, GPA utilizes the graph-based information propagation to obtain the prototype representation of each category.
				However, these two approaches heavily rely on the region proposal step to generate instance-level features, which may be noisy due to the error accumulation problem. More importantly, they neglect the valuable low-level features and the rich interactions among objects of different categories. Our better performance over them (from 38.6\% and 39.5\% to 40.8\%) further reveals the importance of relational reasoning for DAOD tasks.
				
				\item In essence, the classical one-vs-one alignment (such as adversarial training) can be seen as the simplest case of relational reasoning.
				Most conventional domain adaptation methods assume that perfect alignment equals to precise knowledge transfer, while the many-vs-many relations between different entities are ignored. Moreover, alignment-based approaches naturally neglect the intra-domain relations as no entities can be explicitly aligned within each domain.  
				In this regard, our work gives a hint to bridge the gap between alignment-based and relational reasoning based DAOD.
			\end{enumerate}
			
			\textbf{Synthetic-to-Real.} Table~\ref{table5} presents the results on adaptation from Sim10K to Cityscapes. 
			From the table, we can observe that the proposed FGRR algorithm significantly improves the baseline methods.
			The domain disparity of this adaptation task mainly stems from the difference of image style, making it more sophisticated than normal-to-foggy adaptation where only weather changes. In that case, we argue that FGRR explicitly extract the source and target foreground pixels and regions to promote the positive interactions of foregrounds via relational reasoning procedure as well as mitigate the negative influence of background noises.
			By comparison, prior efforts, which aim at either holistic image alignment or local region alignment, cannot ensure the precise knowledge transfer among foreground objects across domains.  
			In particular, methods introducing a FPN architecture would significantly improve over the baseline models. The justification is that this adaptation task focuses on detecting cars which have large scale variances across different scenes, naturally fitting the feature pyramid property of FPN architecture. 
			However, most existing works do not integrate the FPN into their detection backbone, and it may be non-trivial to combine them especially when the DAOD methods have adaptation modules within the backbone networks. To facilitate a fair comparison, we follow the mainstream settings.

			\textbf{Cross-Site Mammogram Mass Detection.} Table~\ref{table6} provides the experimental results on mammogram mass detection, which targets on testing the scalability and efficacy of the proposed FGRR on real-world applications. Cross-site mammogram mass detection substantially differs from adaptation based on nature images due to the particular challenges of sparse distribution and tiny size of lesions, difference of imaging quality as well as existence of hard mimics.
			According to the table, we can see that the proposed FGRR outperforms all comparison methods by a large margin on this challenging adaptation task (from 38.4\% to 43.9\%), clearly demonstrating the effect of our approach on highlighting foreground pixels versus regions and explicitly modeling foreground interactions on both intra- and inter-domains. Moreover, such practical application reveal the potential of DAOD model for improving complex problems.
			
			\begin{center}
				\begin{table*}[htb]
					\caption{Results on PASCAL VOC $\rightarrow$ Clipart Dataset (\%).}\label{table7}
					\centering
					\qquad 
					\setlength\tabcolsep{1.5pt}
					\begin{tabular}{c|cccccccccccccccccccc|cc|a}
						\toprule
						Methods & \rotatebox[origin=c]{90}{aero} & \rotatebox[origin=c]{90}{bcycle} & \rotatebox[origin=c]{90}{bird} & \rotatebox[origin=c]{90}{boat} & \rotatebox[origin=c]{90}{bottle} & \rotatebox[origin=c]{90}{bus} & \rotatebox[origin=c]{90}{car} & \rotatebox[origin=c]{90}{cat} & \rotatebox[origin=c]{90}{chair} & \rotatebox[origin=c]{90}{cow} & \rotatebox[origin=c]{90}{table} & \rotatebox[origin=c]{90}{dog} & \rotatebox[origin=c]{90}{hrs} & \rotatebox[origin=c]{90}{bike} & \rotatebox[origin=c]{90}{prsn} & \rotatebox[origin=c]{90}{plnt} & \rotatebox[origin=c]{90}{sheep} & \rotatebox[origin=c]{90}{sofa} & \rotatebox[origin=c]{90}{train} & \rotatebox[origin=c]{90}{tv} & mAP & mAP* & Gain \\
						\hline
						DANN~\cite{ganin2016domain} & 24.1 & 52.6 & 27.5 & 18.5 & 20.3 & 59.3 & 37.4 & 3.8 & 35.1 & 32.6 & 23.9 & 13.8 & 22.5 & 50.9 & 49.9 & 36.3 & 11.6 & 31.3 & 48.0 & 35.8 & 31.8 & 26.7 & 5.1 \\
						DT+PL w/o label~\cite{inoue2018cross} & 16.8 & 53.7 & 19.7 & \textbf{\color{blush}31.9} & 21.3 & 39.3 & 39.8 & 2.2 & \textbf{\color{blush}42.7} & \textbf{\color{blush}46.3} & 24.5 & 13.0 & \textbf{\color{blush}42.8} & 50.4 & 53.3 & 38.5 & 14.9 & 25.1 & 41.5 & 37.3 & 32.7 & 26.7 & 6.0 \\
						WST~\cite{kim2019self} & 30.8 & 65.5 & 18.7 & 23.0 & 24.9 & 57.5 & 40.2 & 10.9 & 38.0 & 25.9 & \textbf{\color{blush}36.0} & 15.6 & 22.6 & 66.8 & 52.1 & 35.3 & 1.0 & 34.6 & 38.1 & 39.4 & 33.8 & 26.7 & 7.1 \\
						BSR~\cite{kim2019self} & 26.3 & 56.8 & 21.9 & 20.0 & 24.7 & 55.3 & 42.9 & 11.4 & 40.5 & 30.5 & 25.7 & 17.3 & 23.2 & 66.9 & 50.9 & 35.2 & 11.0 & 33.2 & 47.1 & 38.7 & 34.0 & 26.7 & 7.3 \\
						BSR+WST~\cite{kim2019self} & 28.0 & 64.5 & 23.9 & 19.0 & 21.9 & 64.3 & 43.5 & \textbf{\color{blush}16.4} & 42.2 & 25.9 & 30.5 & 7.9 & 25.5 & 67.6 & 54.5 & 36.4 & 10.3 & 31.2 & \textbf{\color{blush}57.4} & 43.5 & 35.7 & 26.7 & 9.0 \\
						I3Net~\cite{chen2021i3net} & 30.0 & 67.0 & \textbf{\color{blush}32.5} & 21.8 & \textbf{\color{blush}29.2} & 62.5 & 41.3 & 11.6 & 37.1 & 39.4 & 27.4 & 19.3 & 25.0 & 67.4 & 55.2 & \textbf{\color{blush}42.9} & \textbf{\color{blush}19.5} & 36.2 & 50.7 & 39.3 & 37.8 & 26.7 & 11.1 \\
						\hline
						DBGL~\cite{chen2021dual} & 23.2 & 65.5 & 30.1 & 18.3 & 24.6 & \textbf{\color{blush}67.6} & \textbf{\color{blush}43.9} & 15.1 & 38.7 & 36.4 & 31.3 & \textbf{\color{blush}20.2} & 25.0 & 74.3 & 55.1 & 38.2 & 12.5 & 41.0 & 49.1 & 43.9 & 37.7 & 26.7 & 11.0 \\
						FGRR~(Ours) & \textbf{\color{blush}33.4} & \textbf{\color{blush}69.5} & 26.4 & 20.8 & 27.4 & 58.1 & 42.3 & 14.2 & 42.0 & 39.7 & 30.4 & 19.6 & 28.7 & \textbf{\color{blush}76.4} & \textbf{\color{blush}56.7} & 40.6 & 8.5 & \textbf{\color{blush}41.3} & 52.1 & \textbf{\color{blush}44.5} & \textbf{\color{blush}38.6} & 26.7 & \textbf{\color{blush}11.9} \\
						\bottomrule
					\end{tabular}
				\end{table*}
			\end{center}
			
			\begin{center}
				\begin{table*}[thb]
					\caption{Ablation of FGRR on six DAOD tasks (\%). 
					}\label{table10}
					\centering
					\begin{tabular}{c|cccccc|c}
						\toprule
						Source Domain & Pascal VOC & Pascal VOC & Pascal VOC & Cityscapes & Sim10k & Public & \multirow{2}*{Avg} \\
						Target Domain & Clipart1k & Watercolor2k & Comic2k & Foggy-Cityscapes & Cityscapes & In-house & \\ 
						\hline
						Source Only & 27.8 & 44.6 & 24.4 & 20.3 & 34.6 & 12.0 & 27.3 \\
						FGRR w/o PRR & 41.0 & 53.6 & 30.2 & 38.6 & 42.5 & 40.0 & 41.0 \\
						FGRR w/o SRR & 41.4 & 53.0 & 30.9 & 37.8 & 41.9 & 39.5 & 40.8 \\
						FGRR w/o IOR & 42.4 & 54.8 & 31.4 & 40.5 & 43.3 & 41.6 & 42.3 \\
						PRR w/o inter & 41.8 & 54.0 & 30.9 & 39.8 & 42.7 & 41.6 & 41.8 \\
						PRR w/o intra & 42.5 & 54.4 & 31.0 & 40.1 & 43.2 & 42.6 & 42.3 \\
						SRR w/o inter & 41.2 & 53.7 & 30.2 & 39.4 & 42.6 & 42.1 & 41.5 \\
						SRR w/o intra & 42.4 & 54.5 & 30.7 & 39.9 & 43.0 & 42.4 &  42.1 \\
						PRR w/o BGL & 41.4 & 53.9 & 31.9 & 39.6 & 42.9 & 43.0 & 42.1 \\
						PRR w/ random link & 39.1 & 52.6 & 28.5 & 37.7 & 40.1 & 39.8 & 39.6 \\
						SRR w/o BGL & 40.9 & 53.3 & 31.4 & 39.4 & 43.0 & 40.4 & 41.4 \\
						SRR w/o CDA & 42.3 & 54.8 & 31.5 & 40.0 & 43.1 & 42.3 & 42.3 \\
						\hline
						FGRR (Full) & 43.3 & 55.7 & 32.7 & 40.8 & 44.5 & 43.9 & 43.5 \\
						\bottomrule
					\end{tabular}
				\end{table*}
			\end{center}
			
			\begin{center}
				\begin{table}[!t]
					\caption{Results on Pascal VOC $\rightarrow$ Watercolor2k (\%).}\label{table8}
					\centering
					\setlength\tabcolsep{3.5pt}
					\scalebox{0.9}{
						\begin{tabular}{c|cccccc|cc|a}
							\toprule
							Methods & bike & bird & car & cat & dog & person & mAP & mAP* & Gain \\
							\hline
							DANN~\cite{ganin2016domain} & 73.4 & 41.0 & 32.4 & 28.6 & 22.1 & 51.4 & 41.5 & 47.1 & -5.6 \\
							BSR~\cite{kim2019self} & 82.8 & 43.2 & 49.8 & 29.6 & 27.6 & 58.4 & 48.6 & 47.1 & 1.5 \\
							WST~\cite{kim2019self} & 77.8 & 48.0 & 45.2 & 30.4 & 29.5 & 64.2 & 49.2 & 47.1 & 2.1 \\
							BSR+WST~\cite{kim2019self} & 75.6 & 45.8 & 49.3 & 34.1 & 30.3 & 64.1 & 49.9 & 47.1 & 2.8 \\
							I3Net~\cite{chen2021i3net} & 81.1 & \textbf{\color{blush}49.3} & 46.2 & 35.0 & 31.9 & 65.7 & 51.5 & 47.1 & 4.4 \\
							\hline
							DBGL~\cite{chen2021dual} & \textbf{\color{blush}84.0} & 46.7 & 45.5 & 36.2 & \textbf{\color{blush}35.7} & 63.7 & 52.0 & 47.1 & 4.9 \\
							FGRR (Ours) & 81.1 & 41.4 & \textbf{\color{blush}53.5} & \textbf{\color{blush}40.7} & 35.0 & \textbf{\color{blush}67.5} & \textbf{\color{blush}53.2} & 47.1 & \textbf{\color{blush}6.1} \\
							\bottomrule
					\end{tabular}}
				\end{table}
			\end{center}
			
			\begin{center}
				\begin{table}[!t]
					\caption{Results on Pascal VOC $\rightarrow$ Comic2k (\%).}\label{table9}
					\centering
					\setlength\tabcolsep{3.5pt}
					\scalebox{0.9}{
						\begin{tabular}{c|cccccc|cc|a}
							\toprule
							Methods & bike & bird & car & cat & dog & person & mAP & mAP* & Gain \\
							\hline
							DANN~\cite{ganin2016domain} & 33.3 & 11.3 & 19.7 & 13.4 & 19.6 & 37.4 & 22.5 & 21.9 & 0.6 \\
							BSR~\cite{kim2019self} & 45.2 & 15.8 & 26.3 & 9.9 & 15.8 & 39.7 & 25.5 & 21.9 & 3.6 \\
							WST~\cite{kim2019self} & 45.7 & 9.3 & 30.4 & 9.1 & 10.9 & 46.9 & 25.4 & 21.9 & 3.5 \\
							BSR+WST~\cite{kim2019self} & \textbf{\color{blush}50.6} & 13.6 & 31.0 & 7.5 & 16.4 & 41.4 & 26.8 & 21.9 & 4.9 \\
							I3Net~\cite{chen2021i3net} & 47.5 & \textbf{\color{blush}19.9} & \textbf{\color{blush}33.2} & 11.4 & 19.4 & 49.1 & 30.1 & 21.9 & 8.2 \\
							\hline
							DBGL~\cite{chen2021dual} & 45.4 & 15.9 & 24.8 & 11.5 & \textbf{\color{blush}29.4} & 55.1 & 30.4 & 21.9 & 8.5 \\
							FGRR~(Ours) & 49.8 & 18.0 & 29.3 & \textbf{\color{blush}13.8} & 25.5 & \textbf{\color{blush}55.2} & \textbf{\color{blush}31.9} & 21.9 & \textbf{\color{blush}10.0} \\
							\bottomrule
					\end{tabular}}
				\end{table}
			\end{center}
			\vspace{-2.5cm}
			\subsection{FGRR for Adapting One-Stage Object Detector}
			\label{sec:5-4}
			Most existing DAOD approaches are built on the top of Faster R-CNN detection framework and, consequently, how to adapting one-stage object detectors is yet to be thoroughly studied. In this section, we generalize the proposed FGRR algorithm for adapting one-stage detectors. Following the only two works~\cite{kim2019self,chen2021i3net} that have tackled the task of adapting one-stage detectors, we adopt SSD300~\cite{liu2016ssd} framework with VGG-16~\cite{simonyan2014very} architectures. To facilitate a fair comparison, we also follow them~\cite{kim2019self,chen2021i3net} to conduct experiments on three adaptation tasks, \emph{i.e.,} PASCAL VOC $\rightarrow$ Clipart, PASCAL VOC $\rightarrow$ Watercolor2k, and Pascal VOC $\rightarrow$ Comic2k. In experiments, we resize the input images to 300~$\times$~300 and conduct data augmentations by following~\cite{liu2016ssd,kim2019self}. The batch size is set as 16 to fit the GPU memory.
			
			The adaptation results are presented in Table~\ref{table7}, Table~\ref{table8}, and Table~\ref{table9} respectively.   
			The proposed FGRR outperforms all the baseline methods in terms of mAP. 
			It is noteworthy that FGRR improves over the results of DBGL by +1.2\% on average (37.7\% $\rightarrow$ 38.6\%, 52.0\% $\rightarrow$ 53.2\%, and 30.4\% $\rightarrow$ 31.9\%).
			DANN stands for the method that embeds the vanilla adversarial feature adaptation mechanism into the detection pipeline.
			From the result, we can see that simply incorporating the alignment module with one-stage detectors achieves inferior performance. For example, in the task of PASCAL VOC $\rightarrow$ Clipart, the Source Only model unexpectedly outperforms DANN by a large margin (from 41.5\% to 47.1\%), implying that DANN is prone to cause negative transfer in this case. By comparison, vanilla adversarial adaptation module could substantially improve the Source Only baseline when adapting Faster R-CNN (such as DA-Faster~\cite{chen2018domain}). This phenomenon indicates that it is non-trivial to extend a DAOD algorithm to adapting one-stage detectors. WST+BSR~\cite{kim2019self} is tailored for SSD framework and thus demonstrates strong efficacy.
			In particular, I3Net implicitly learn instance-invariant features by changing the emphasis of adaptation from holistic to local, \emph{i.e.,} attending to discriminative regions versus samples and aligning semantic-level representations. However, joining the practice of prior two-stage DAOD methods, this work focuses on one-to-one matching and neglects the significance of exploring many-to-many relationships.
			In this regard, the proposed FGRR not only highlights the foreground pixels and regions without introducing any special modules but also tackles the intra- and inter-domain relational reasoning problems via graph-based structures in the context of DAOD.  
			The better performance achieved by our method verifies this point.
			
			\begin{figure}[!t]
				\centering
				\subfigure[Target: Foggy-Cityscapes]{\includegraphics[width=0.23\textwidth]{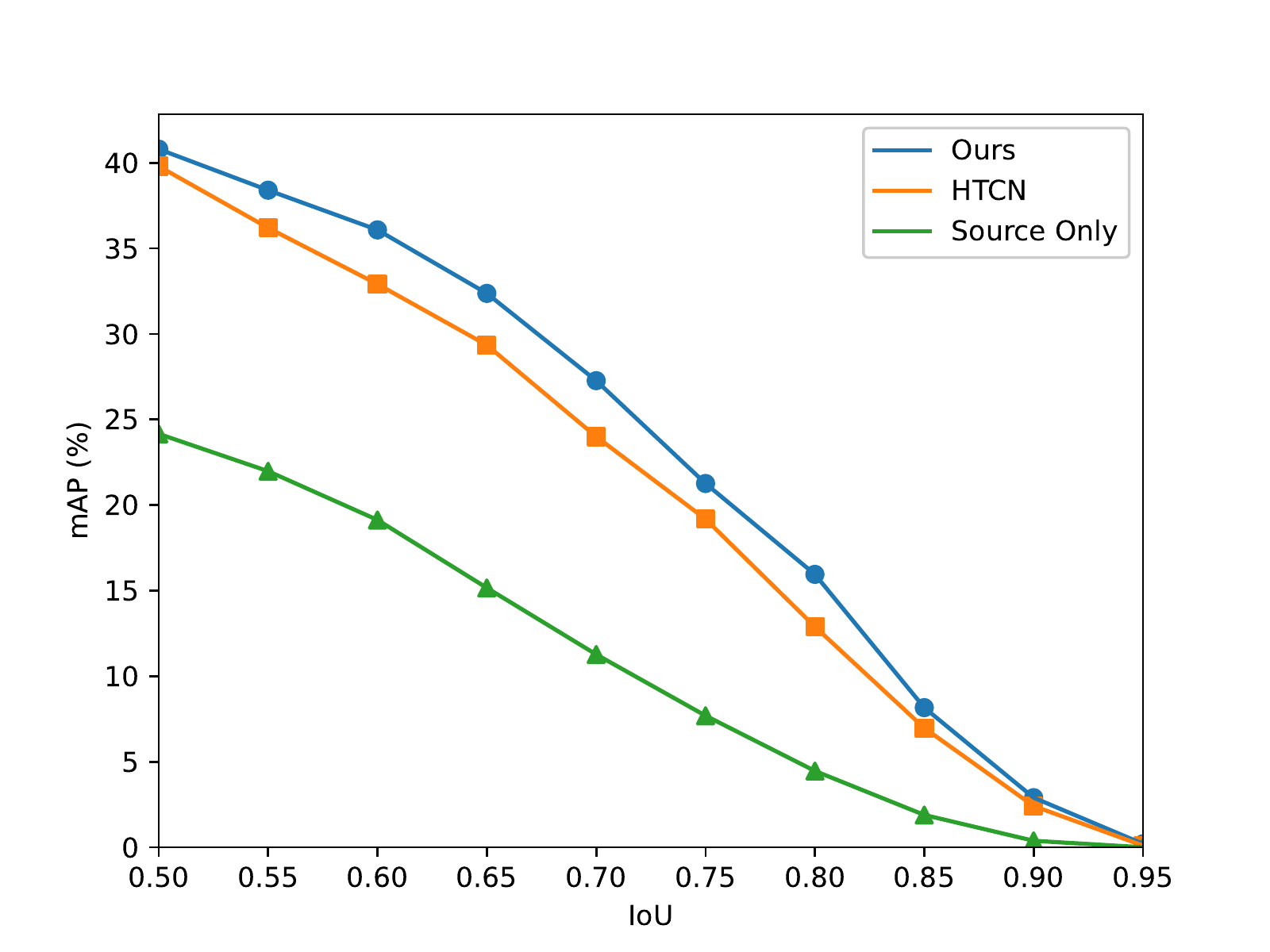}}
				\subfigure[Target: Clipart1k]{\includegraphics[width=0.23\textwidth]{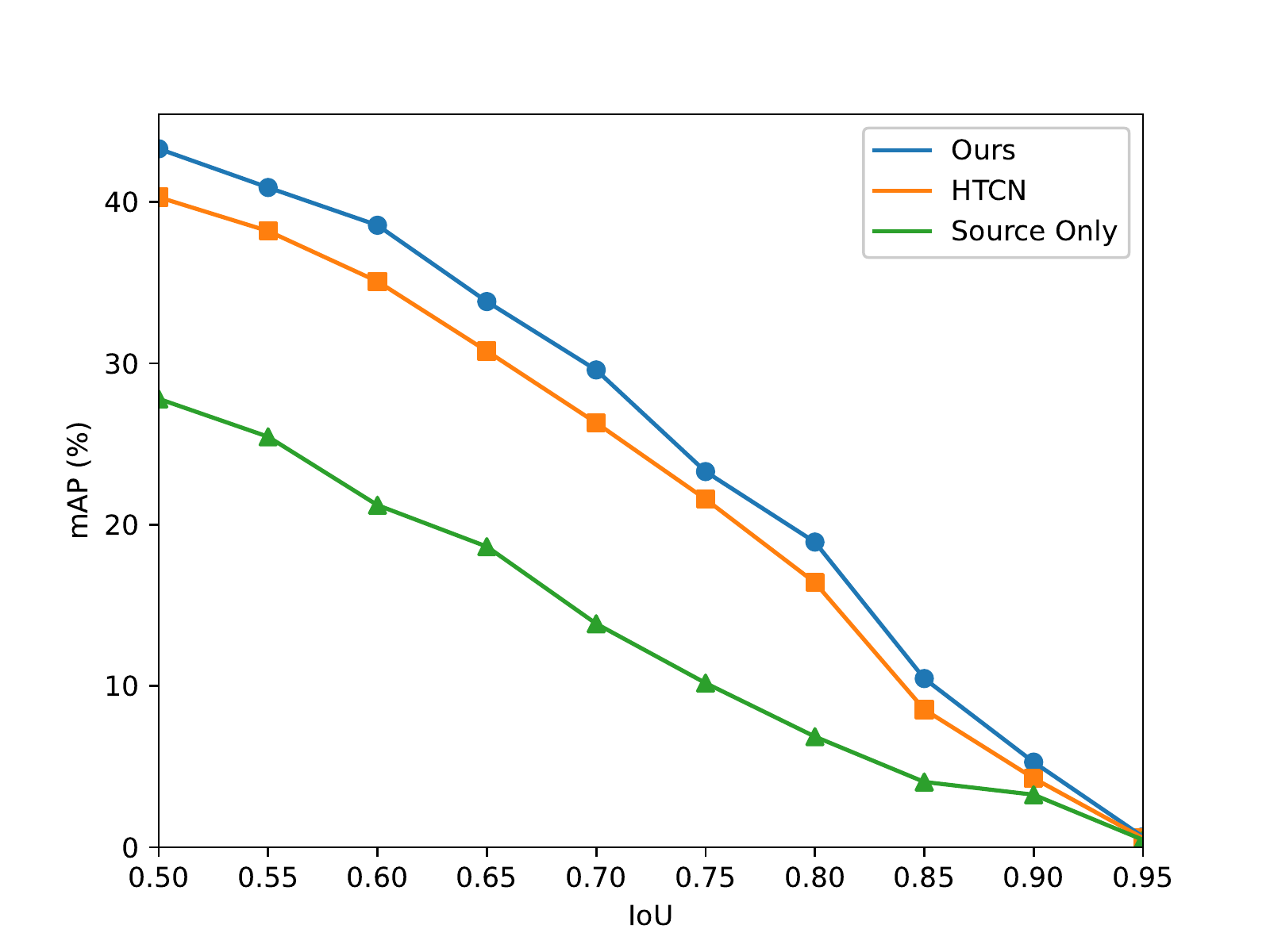}}
				\caption{performance with the variation of IOU thresholds on adaptation tasks Cityscapes~$\rightarrow$~Foggy-Cityscapes and Pascal VOC~$\rightarrow$~Clipart1k.}
				\label{fig:iou}
			\end{figure}
			
			\begin{figure*}[!t]
				\centering
				\subfigure[Source Only]{\includegraphics[width=0.3\textwidth]{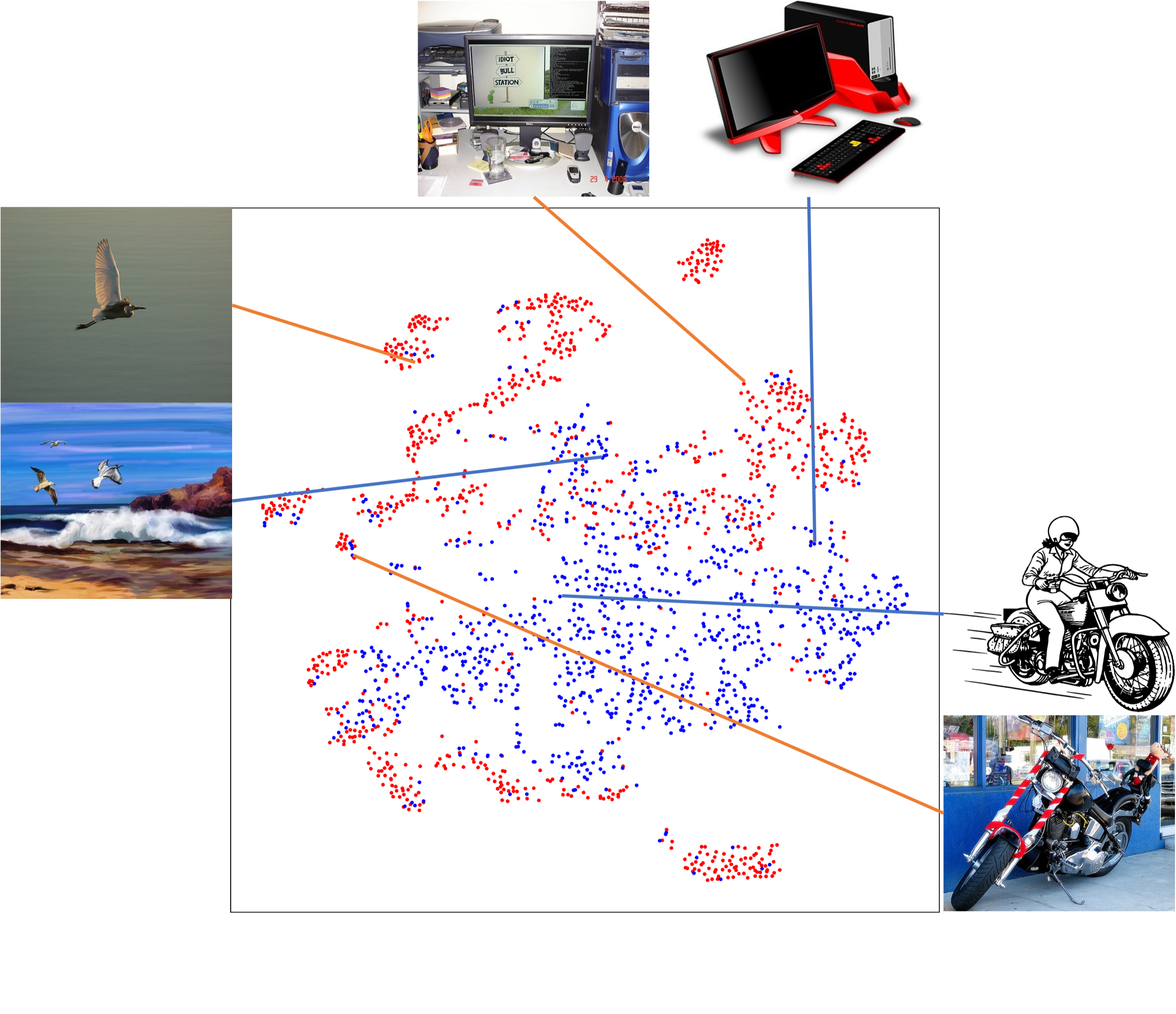}}
				\subfigure[HTCN]{\includegraphics[width=0.3\textwidth]{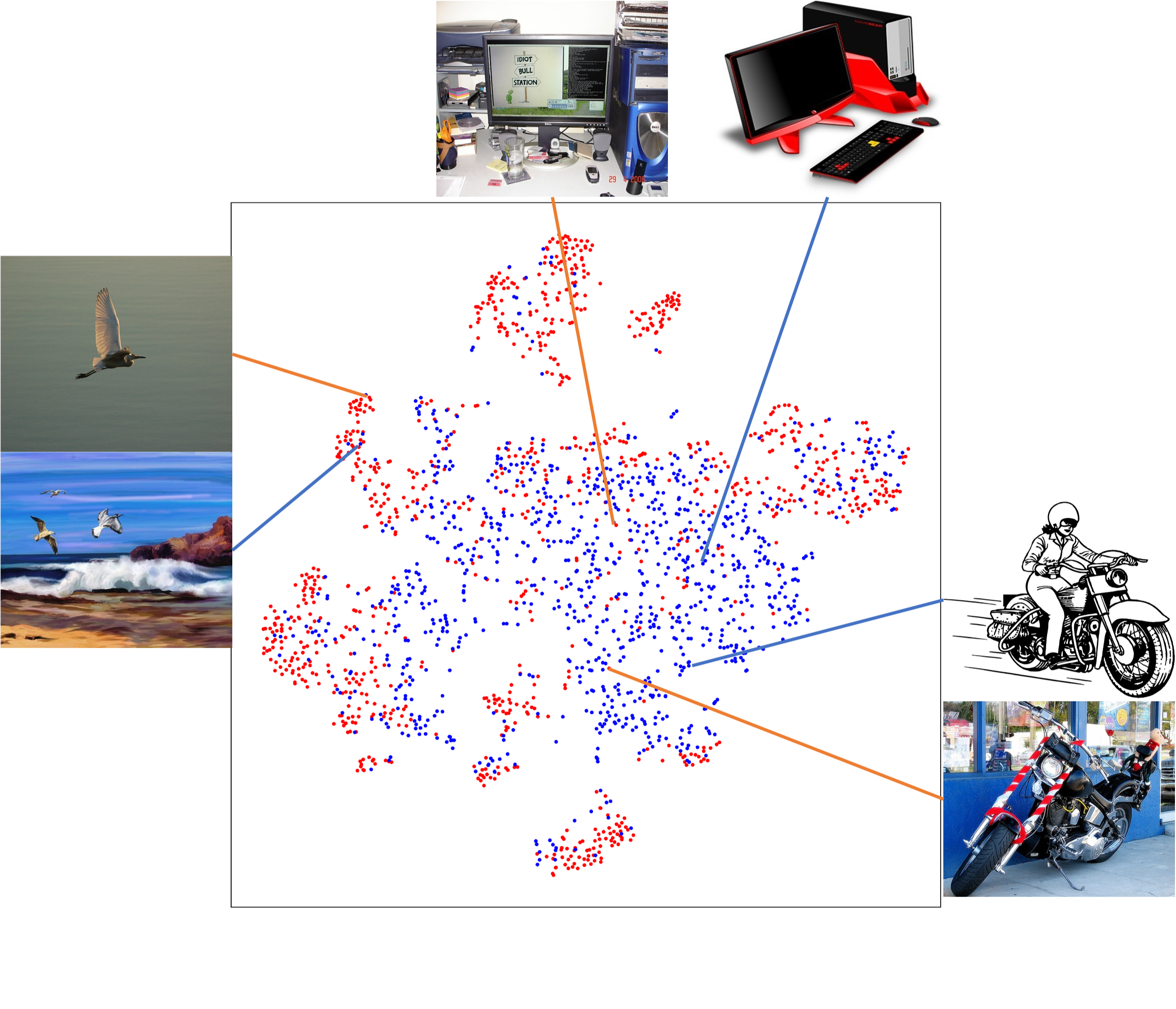}}
				\subfigure[FGRR (Ours)]{\includegraphics[width=0.3\textwidth]{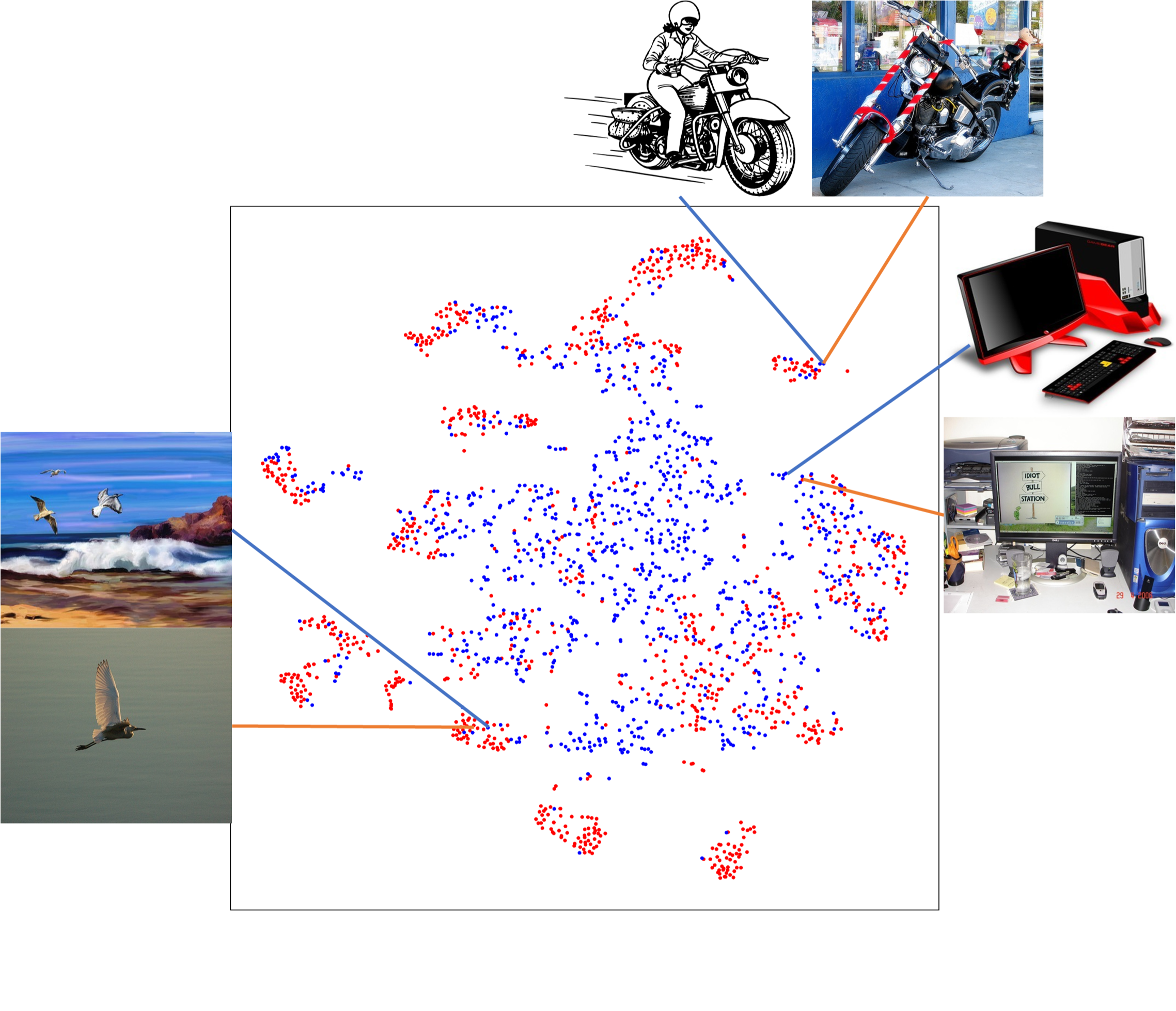}}
				\caption{The t-SNE visualization of extracted features on adaptation task Pascal VOC~$\rightarrow$~Clipart1k. Red: source features, Blue: target features.
					Images with orange lines are from the source domain and images with blue lines are from target domain.}
				\label{fig:tsne}
			\end{figure*}
			
			\begin{figure*}[!h]
				\centering
				\includegraphics[width=0.8\textwidth]{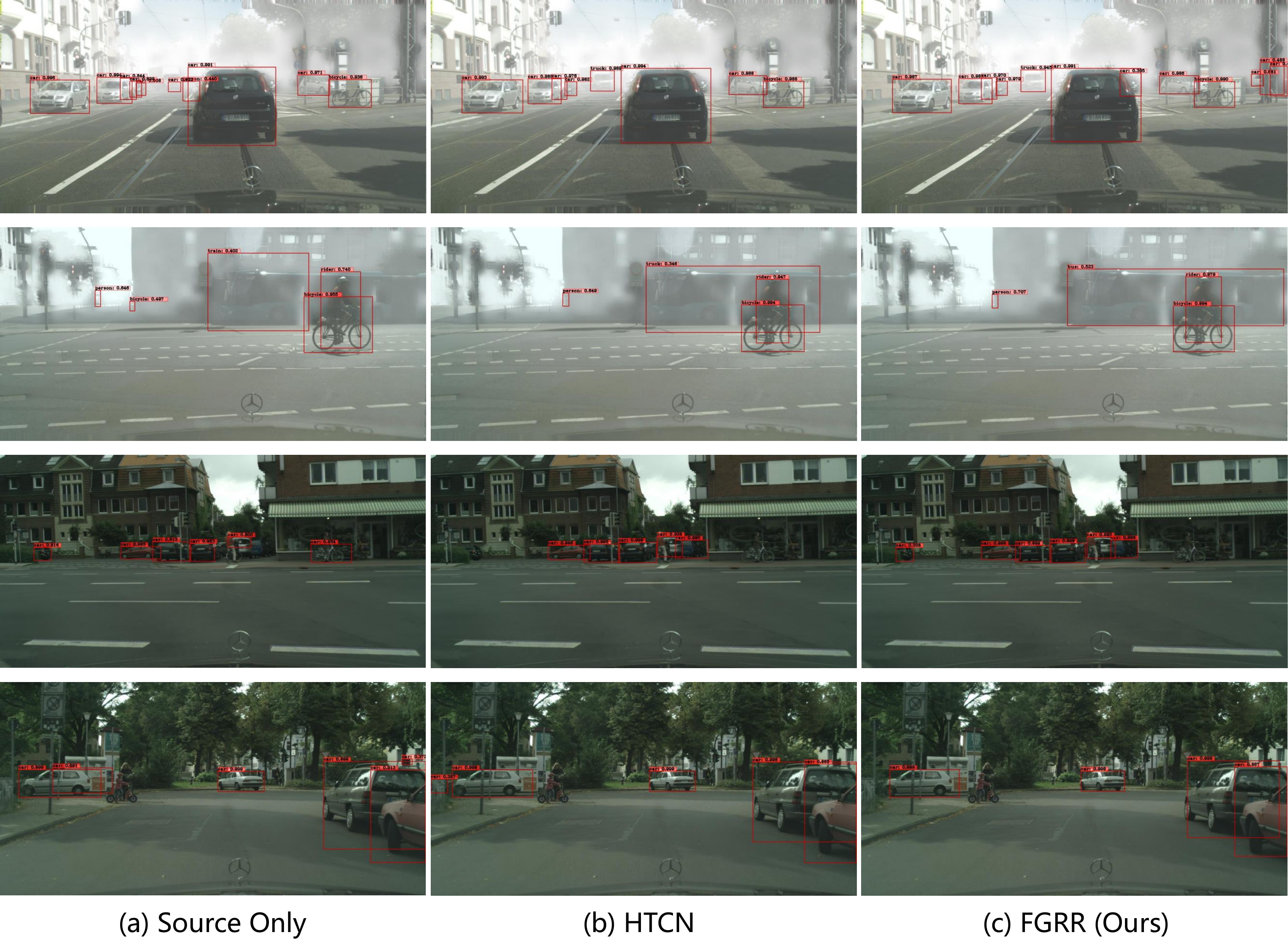}
				\vspace{-0.2cm}
				\caption{Detection results on the target domain. First and second rows: Cityscapes~$\rightarrow$~Foggy-Cityscapes. Third and fourth rows: Sim10K~$\rightarrow$~Cityscapes.}\label{fig:cs}
			\end{figure*}
			
			\begin{figure*}[htb]
				\centering
				\includegraphics[width=0.8\textwidth]{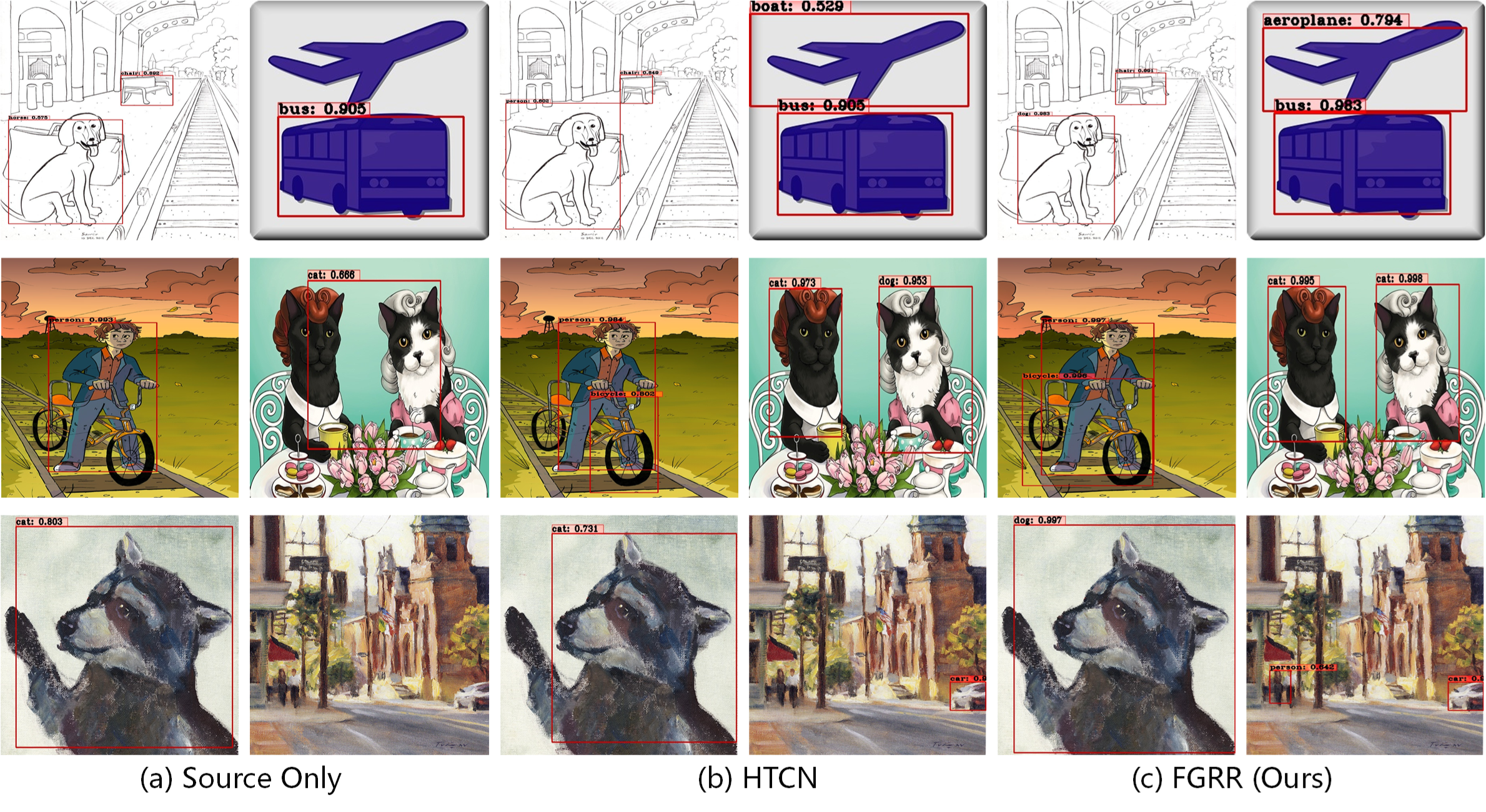}
				\vspace{-0.2cm}
				\caption{Detection results on the target domain. First row: Pascal VOC~$\rightarrow$~Clipart1k. Second row: Pascal VOC~$\rightarrow$~Watercolor2k. Third row: Pascal VOC~$\rightarrow$~Comic2k.}\label{fig:voc}
			\end{figure*}
			
			\begin{figure*}[htb]
				\centering
				\includegraphics[width=0.8\textwidth]{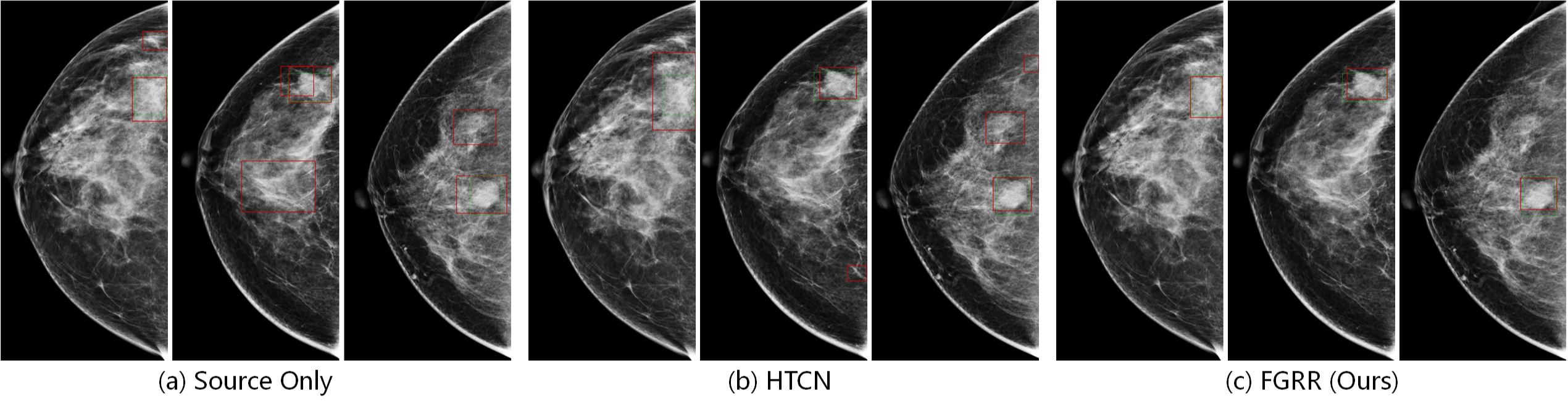}
				\vspace{-0.2cm}
				\caption{Detection results on In-house dataset. Bounding boxes with green color denote the ground-truth, and with red color denote the prediction results.}\label{fig:breast}
			\end{figure*}
			
			\subsection{Further Empirical Analysis}
			\subsubsection{Ablation Study} 
			We investigate the individual effects of each proposed module by performing in-depth and complete ablation studies. 
			We evaluate several variants of FGRR: (1) \emph{FGRR w/o PRR}, \emph{FGRR w/o SRR}, and \emph{FGRR w/o IOR}: remove the pixel-level and semantic relational reasoning modules as well as the image-level object-aware reweighting module from the full FGRR model respectively. 
			(2) \emph{PRR w/o inter} and \emph{PRR w/o intra}: remove the inter- and intra-domain reasoning modules from the PRR respectively. 
			\emph{SRR w/o inter} and \emph{SRR w/o intra}: remove the inter- and intra-domain reasoning modules from the SRR respectively. 
			(3) \emph{PRR w/o BGL}: directly learn the cross-domain one-vs-one correspondence based on searched foreground pixel pairs without introducing bipartite graph. 
			\emph{PRR w/ random link}: randomly select pixels to construct bipartite graph. 
			(4) \emph{SRR w/o BGL}: directly learn the cross-domain one-vs-one correspondence based on foreground proposals without introducing bipartite graph..
			\emph{SRR w/o CDA}: remove the category-aware domain alignment module from SRR. 
			The results of ablation study are summarized in Table~\ref{table10}.
			From the table, we have the following observations. 
			(1) Respectively removing the PRR and SRR components from the full FGRR model will lead to substantial performance degeneration, clearly revealing their individual effects and the complementary effect between intra- and inter-domain relational reasoning. 
			(2) The results of FGRR w/o IOR verify that estimating the object-level variations based on image-level features can harmonize the adaptation process and thus provide better knowledge transfer.
			(3) Inter-domain and intra-domain relational reasoning modules make almost equal contributions to the final performance, highlighting the necessity of considering them in a unified framework.
			(4) The performances of PRR w/o BGL and SRR w/o BGL drop significantly compared to the full FGRR model, demonstrating the importance of performing relational reasoning compared to conventional local and global alignment approaches. 
			(5) The results of PRR w/ random link reveal that randomly linking graph node will bring in a number of noisy connections and thereby cause significant performance degeneration, showing the superiority of graph construction strategies in our method. 
			
			\subsubsection{Influence of IoU threshold} 
			Considering that the IoU threshold is an important factor that affects the detection results, we study the adaptation performance of three models (\emph{i.e.,} Source Only, HTCN~\cite{chen2020harmonizing}, and FGRR) with different IoU threshold on Cityscapes~$\rightarrow$~Foggy-Cityscapes and Pascal VOC~$\rightarrow$~Clipart1k.
			The results are reported in Figure~\ref{fig:iou}, where we can see that 1) the mAP gradually decreases with the increasing of IoU threshold and approaches zero at the end, and 2) the proposed FGRR consistently outperforms the comparison models on different threshold, revealing the effectiveness of our FGRR on providing robust and precise bounding boxes prediction.  
			
			\subsubsection{Feature Visualization}
			Figure~\ref{fig:tsne} visualizes the image-level features generated by Source Only, HTCN and our FGRR using t-SNE algorithm on adaptation task Pascal VOC~$\rightarrow$~Clipart1k. As can be seen, for our FGRR model, images that have similar semantic information will get much closer in the feature space compared to the other models. It is noteworthy that when compared to Source Only and HTCN, FGRR does not match the source and target features more compact, but improves the adaptation performance by a large margin in terms of mAP, implying that globally matching features cannot ensure the fine-grained knowledge transfer. This result also verifies the hypothesis in~\cite{saito2019strong} that weak global alignment helps the adaptation of object detectors.  
			
			\subsubsection{Qualitative Detection Results} 
			Figure~\ref{fig:cs}, Figure~\ref{fig:voc}, and Figure~\ref{fig:breast} provide some detection results on the target domain based on different DAOD methods, \emph{i.e.,} Source Only, HTCN~\cite{chen2020harmonizing}, and FGRR. 
			From the figures, we can see that the proposed FGRR is significantly better than the compared methods in terms of giving accurate boxes regression and object classification. More specifically, FGRR is capable of precisely detecting those sample-scarce or/and hard-to-classify categories, obscured or/and tiny foreground objects, as well as largely reducing the false positive results.

\section{Conclusion}
\label{sec:conclusion}
In this paper, we delve into the relational reasoning problem for domain adaptive object detection. A Foreground-aware Graph-based Relational Reasoning (FGRR) framework is introduced to explicitly endow the detection models with the capability of reasoning over relations between foreground objects on both pixel-level and semantic-level. The proposed FGRR first identifies the foreground pixels and regions to represent graph nodes. Then, the graph edges are constructed by regularizing the affinity between nodes, and finally, the inter- and intra-domain relations is learned separately via bipartite graph learning and graph attention mechanisms. Experiments on four DAOD benchmarks validated the effectiveness of the proposed FGRR.

\bibliographystyle{IEEEtran}
\bibliography{IEEEabrv,main}

\begin{IEEEbiography}[{\includegraphics[width=1in,height=1.25in,clip,keepaspectratio]{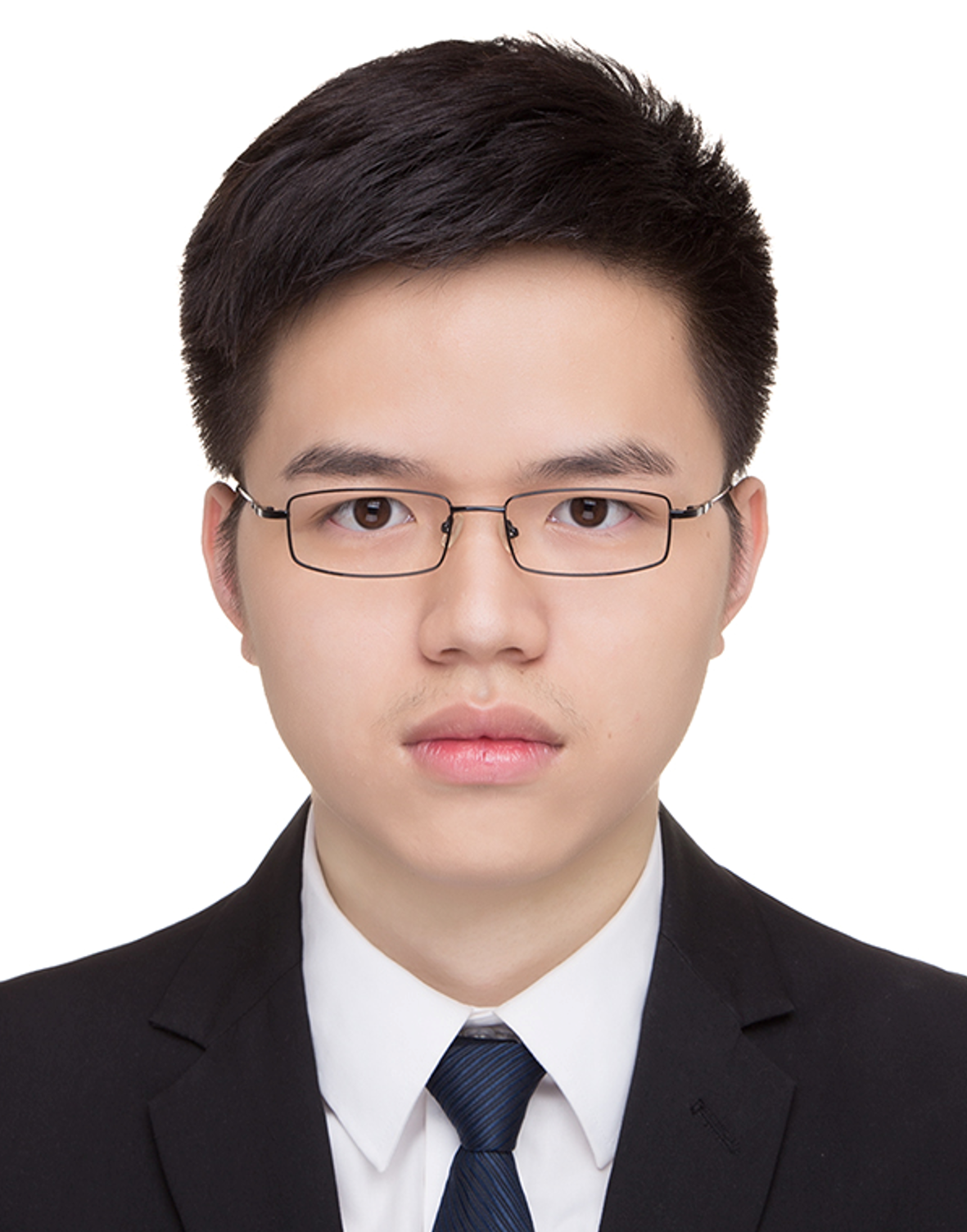}}]{Chaoqi Chen}
	received the B.S. and M.E. Degrees from Xiamen University, China, in 2017 and 2020, respectively. He is currently pursuing the Ph.D. degree with the Department of Computer Science, The University of Hong Kong. His research interest spans on machine learning and computer vision, with special interests in transfer learning and visual reasoning.
\end{IEEEbiography}

\begin{IEEEbiography}[{\includegraphics[width=1in,height=1.25in,clip,keepaspectratio]{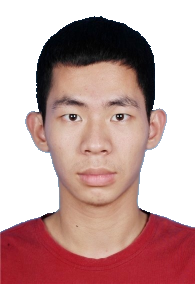}}]{Jiongcheng Li}
	received the B.S. degree from Shenzhen University, China, in 2020. He is currently pursuing the M.S. degree with the Department of Information and Communication Engineering, Xiamen University. His research interests include computer vision and machine learning.
\end{IEEEbiography}

\begin{IEEEbiography}[{\includegraphics[width=1in,height=1.25in,clip,keepaspectratio]{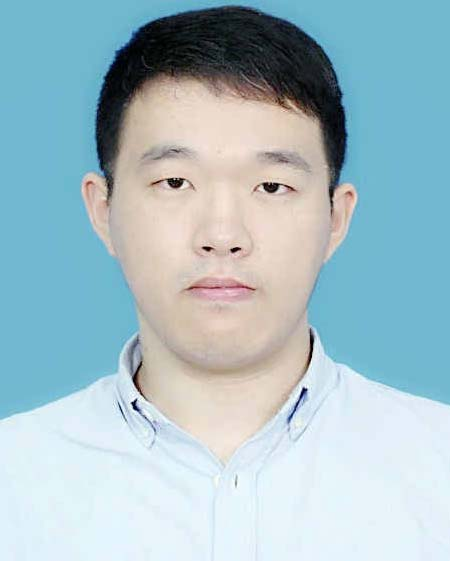}}]{Hong-Yu Zhou}
	received the B.S. degree from Wuhan University, China, in 2015, and the M.S. degree from the Department of Computer Science
	and Technology, Nanjing University, China, in 2018. He is currently pursuing the Ph.D. degree with the Department of Computer Science, The University of Hong Kong. His research interests include computer vision and machine learning.
\end{IEEEbiography}

\begin{IEEEbiography}[{\includegraphics[width=1in,height=1.25in,clip,keepaspectratio]{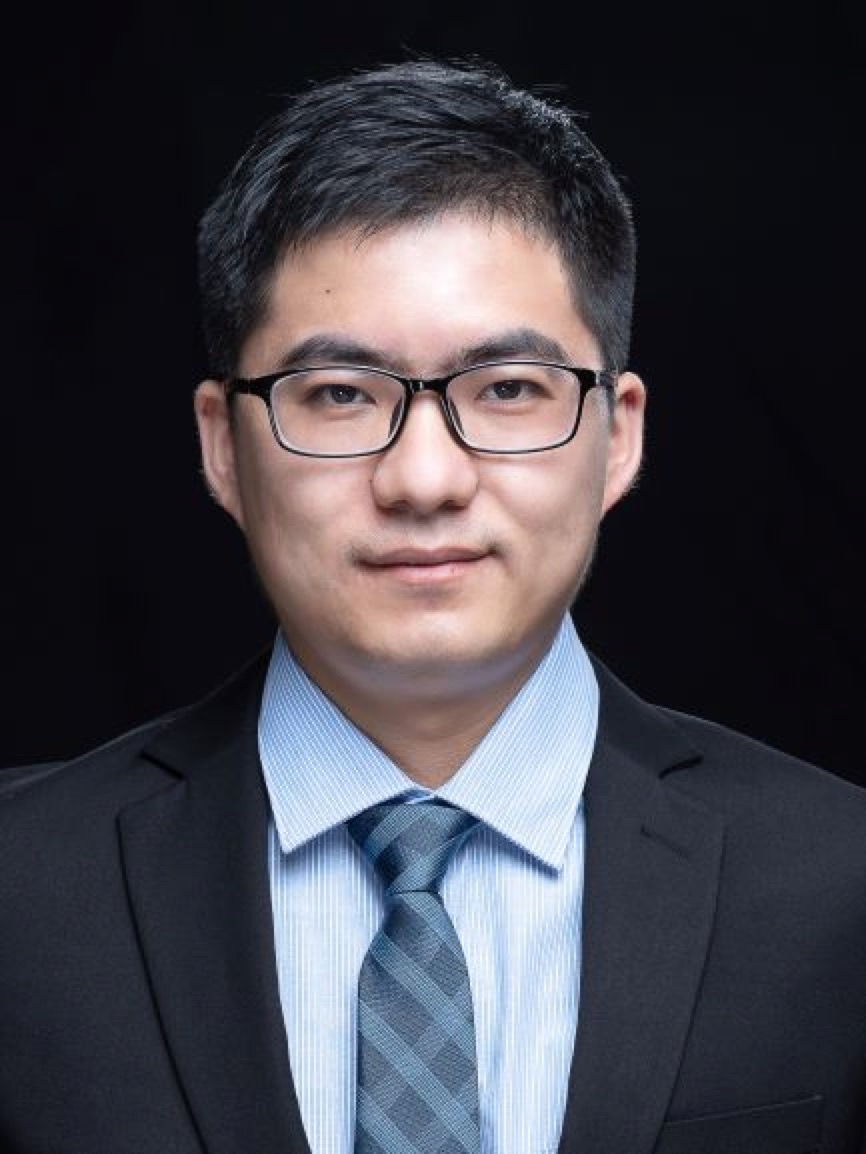}}]{Xiaoguang Han}
	is now an Assistant Professor at The Chinese University of Hong Kong, Shenzhen. He received his Ph.D. degree in computer science from The University of Hong Kong (2013-2017), his M.S. degree in applied mathematics from Zhejiang University (2009-2011) and his B.S. degree in math from Nanjing University of Aeronautics and Astronautics. He also spent 2 years (2011-2013) in City University of Hong Kong as a research associate. His research interests include computer vision, compute graphics, human-computer interaction, medical image analysis, and machine learning.
\end{IEEEbiography}

\begin{IEEEbiography}[{\includegraphics[width=1in,height=1.25in,clip,keepaspectratio]{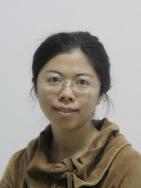}}]{Yue Huang} 
	received the B.S. degree from Xiamen University, Xiamen, China, in 2005, and the Ph.D. degree from Tsinghua University, Beijing, China, in 2010. She was a Visiting Scholar with Carnegie Mellon University from 2015 to 2016. She is currently an Associate Professor with the Department of Information and Communication Engineering, School of Informatics, Xiamen University. Her research interests include machine learning and image processing.
\end{IEEEbiography}

\begin{IEEEbiography}[{\includegraphics[width=1in,height=1.25in,clip,keepaspectratio]{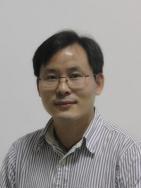}}]{Xinghao Ding} 
	received the B.S. and Ph.D. degrees from the Hefei University of Technology, Hefei, in 1998 and 2003, respectively. From September 2009 to March 2011, he was a Post-Doctoral Researcher with the Department of Electrical and Computer Engineering, Pratt School of Engineering, Duke University, Durham, NC, USA. Since 2011, he has been a Professor with the Department of Information and Communication Engineering, School of Informatics, Xiamen University, Xiamen, China. His research interests include image processing and machine learning.
\end{IEEEbiography}

\begin{IEEEbiography}[{\includegraphics[width=1in,height=1.25in,clip,keepaspectratio]{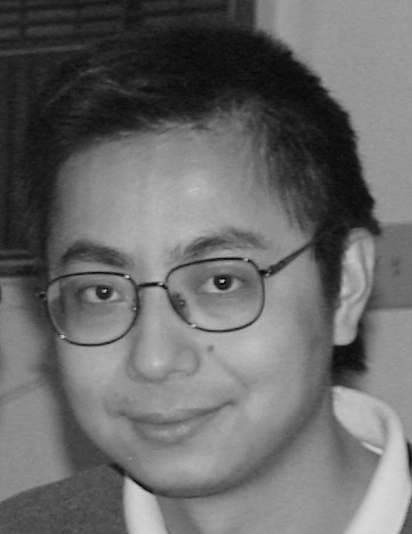}}]{Yizhou Yu}
	(M’10, SM’12, F’19) received the PhD degree from University of California at Berkeley in 2000. He is a professor at The University of Hong Kong, and was a faculty member at University of Illinois at Urbana-Champaign for twelve years. He is a recipient of 2002 US National Science Foundation CAREER Award and ACCV 2018 Best Application Paper Award. Prof Yu has served on the editorial board of IET Computer Vision, The Visual Computer, and IEEE Transactions on Visualization and Computer Graphics. He has also served on the program committee of many leading international conferences, including CVPR, ICCV, and SIGGRAPH. His current research interests include computer vision, deep learning, AI for medicine, and geometric computing.
\end{IEEEbiography}

\end{document}

%% file: math_commands.tex
\usepackage{amsmath,amsfonts,bm}

\def\eqref#1{equation~\ref{#1}}

\def\1{\bm{1}}

\def\va{{\bm{a}}}

\def\mA{{\bm{A}}}

\def\mW{{\bm{W}}}
\def\mX{{\bm{X}}}

\DeclareMathAlphabet{\mathsfit}{\encodingdefault}{\sfdefault}{m}{sl}
\SetMathAlphabet{\mathsfit}{bold}{\encodingdefault}{\sfdefault}{bx}{n}

